\documentclass[fleqn,10pt]{wlscirep}
\usepackage[utf8]{inputenc}
\usepackage[T1]{fontenc}

\usepackage[title]{appendix}%
\usepackage{dirtytalk}
\usepackage{tabularx}

\usepackage{hyperref}
\usepackage{url}
\usepackage{graphicx}
\graphicspath{ {./figures/} }
\usepackage{wrapfig}
\usepackage{lipsum}
\usepackage{xcolor}         
\usepackage{listings}
\usepackage{fancyvrb}
\usepackage{wrapfig}
\usepackage{booktabs}
\definecolor{mBlue}{HTML}{003366}

\usepackage[utf8]{inputenc} 
\usepackage[T1]{fontenc}    
\usepackage{url}            
\usepackage{amsfonts}       
\usepackage{nicefrac}       
\usepackage{microtype}      

\title{Playing repeated games with Large Language Models}
\author[1,2,3,*]{Elif Akata}
\author[2]{Lion Schulz}
\author[1,2]{Julian Coda-Forno}
\author[3]{Seong Joon Oh}
\author[3]{Matthias Bethge}
\author[1,2]{Eric Schulz}
\affil[1]{Institute for Human-Centered AI, Helmholtz Munich, Oberschleißheim, Germany}
\affil[2]{Max Planck Institute for Biological Cybernetics, T\"ubingen, Germany}
\affil[3]{University of T\"ubingen, T\"ubingen, Germany}
\affil[*]{elif.akata@uni-tuebingen.de}

\begin{abstract}
LLMs are increasingly used in applications where they interact with humans and other agents. We propose to use behavioural game theory to study LLM's cooperation and coordination behaviour. We let different LLMs play finitely repeated $2\times2$ games with each other, with human-like strategies, and actual human players. Our results show that LLMs perform particularly well at self-interested games like the iterated Prisoner's Dilemma family. However, they behave sub-optimally in games that require coordination, like the Battle of the Sexes. We verify that these behavioural signatures are stable across robustness checks. We additionally show how GPT-4's behaviour can be modulated by providing additional information about its opponent and by using a “social chain-of-thought” (SCoT) strategy. This also leads to better scores and more successful coordination when interacting with human players. These results enrich our understanding of LLM's social behaviour and pave the way for a behavioural game theory for machines.
\end{abstract}
\begin{document}

\flushbottom
\maketitle

\thispagestyle{empty}
\section*{Introduction}

\begin{figure}[ht!]
    \begin{center}
    \vspace{-10pt}
    \includegraphics[width=\textwidth]{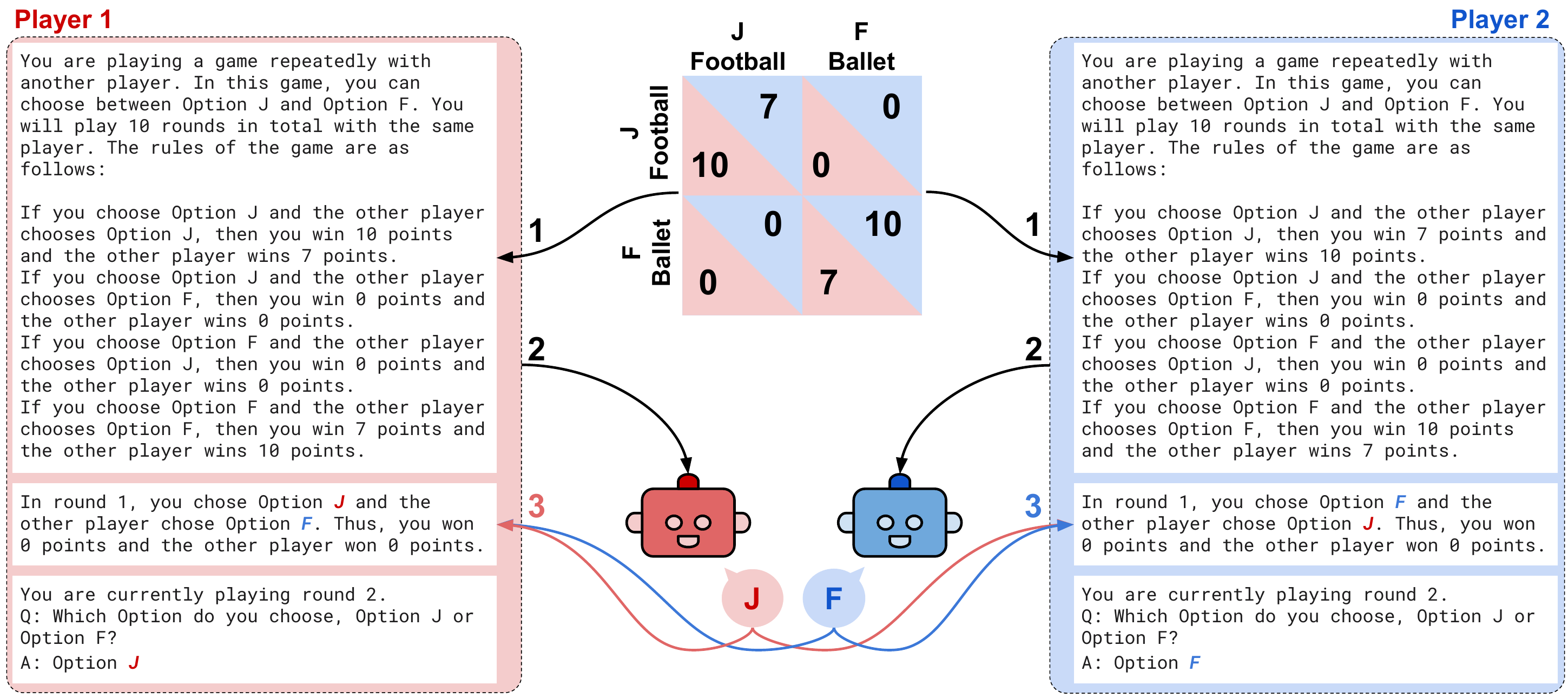}
    \end{center}
    \vspace{-10pt}
    \caption{Playing repeated games in an example game of Battle of the Sexes. In Step (1), the payoff matrix is turned into textual game rules. (2) The game rules, the current game history, and the query are concatenated and passed to LLMs as prompts. (3) In each round, the history for each player is updated with the answers and scores of both players. Steps 2 and 3 are repeated for 10 rounds.}
    \label{fig:overview}
\end{figure}

Large Language Models (LLMs) are deep learning models with billions of parameters trained on huge corpora of text \cite{brants2007large,devlin2018bert,radford2018improving}. While they can generate text that human evaluators struggle to distinguish from text written by other humans \cite{brown2020language}, they have also shown other, emerging abilities \cite{wei2022emergent}. They can, for example, solve analogical reasoning tasks \cite{webb2022emergent}, program web applications \cite{chen2021evaluating}, use tools to solve multiple tasks \cite{bubeck2023sparks}, or adapt their strategies purely in-context \cite{codaforno2023metaincontext}. Because of these abilities and their increasing popularity, LLMs are already transforming our daily lives as they permeate into many applications \cite{bommasani2021opportunities}. This means that LLMs will interact with us and other agents --LLMs or otherwise-- frequently and repeatedly. How do LLMs behave in these repeated social interactions?

Measuring how people behave in repeated interactions, for example, how they cooperate \cite{fudenberg2012slow} and coordinate \cite{mailath2004coordination}, is the subject of a sub-field of behavioural economics called behavioural game theory \cite{camerer2011behavioral}. While traditional game theory assumes that people's strategic decisions are rational, selfish, and focused on utility maximization \cite{fudenberg1991game,VNM}, behavioural game theory has shown that human agents deviate from these principles and, therefore, examines how their decisions are shaped by social preferences, social utility and other psychological factors \cite{camerer1997progress}. Thus, behavioural game theory lends itself well to studying the repeated interactions of diverse agents \cite{henrich2001search,rousseau1998not}, including artificial agents \cite{johnson2022measuring}. 

In this paper, we analyze LLMs' behavioural patterns by letting them play finitely repeated games with full information and against other LLMs, simple, human-like strategies, and actual human players. Finitely repeated games have been engineered to understand how agents should and do behave in interactions over many iterations. We focus on two-player games with two discrete actions, i.e. $2\times2$-games (see Figure \ref{fig:overview} for an overview).

Analyzing LLMs' performance across families of games, we find that they perform well in games that value pure self-interest, especially those from the Prisoner's Dilemma family. However, they underperform in games that involve coordination. Based on this finding, we further focus on games taken from these families and, in particular, on the currently largest LLM: GPT-4 \cite{openai2023gpt4}. In the canonical Prisoner's Dilemma, which assesses how agents cooperate and defect, we find that GPT-4 retaliates repeatedly, even after only having experienced one defection. Because this can indeed be the equilibrium individual-level strategy, GPT-4 is good at these games because it is particularly unforgiving and selfish. In the Battle of the Sexes, which assesses how agents trade-off between their own and their partners' preferences, we however find that GPT-4 does not manage to coordinate with simple, human-like agents, that alternate between options over trials. Thus, GPT-4 is bad at these games because it is uncoordinated. We also verify that these behaviours are not due to an inability to predict the other player's actions, and persist across several robustness checks and changes to the payoff matrices. We point to two ways in which these behaviours can be changed. GPT-4 can be made to act more forgivingly by pointing out that the other player can make mistakes. Moreover, GPT-4 gets better at coordinating with the other player when it is first asked to predict their actions before choosing an action itself, an approach we term social chain-of-thought prompting (SCoT). Finally, we let GPT-4 with and without SCoT-prompting play the canonical Prisoner's Dilemma and the Battle of the Sexes with human players. We find that SCoT-prompting leads to more successful coordination and joint cooperation between participants and LLMs, and makes participants believe more frequently that the other player is human.
    
\section*{Results}\label{sec2}

Using GPT-4, text-davinci-002, text-davinci-003, Claude 2 and Llama 2 70B, we evaluate a range of $2 \times 2$-games. For the analysis of two particular games, we let all the LLMs and human-like strategies play against each other. We focus on LLMs' behaviour in cooperation and coordination games.

\subsection*{Analysing behaviour across families of games}
\label{analysing families}

\begin{table}[t]
    \caption{Performance of all models on 6 families of $2 \times 2$-games. Model score divided by maximum score achievable under ideal conditions. Best performing model is marked in \textbf{bold}.}
    \vspace{-10pt}
    \label{tab:all_games_results}
    \begin{center}
    \begin{tabular}[t]{lccccc}
    \toprule
    Game family & Llama 2 & Claude 2 & davinci-002 & davinci-003 & GPT-4\\
    \midrule
    Second best & 0.486 & 0.735 & 0.473 & 0.692 & \textbf{0.763}\\
    Biased      & 0.632 & 0.794 & 0.629 & 0.761 & \textbf{0.798}\\
    Cyclic      & 0.634 & 0.749 & 0.638 & 0.793 &  \textbf{0.806}\\
    Unfair      & 0.641 & 0.812 & 0.683 & 0.833 & \textbf{0.836}\\
    PD Family   & 0.731 & 0.838 & 0.807 & 0.841 & \textbf{0.871}\\
    Win-win     & 0.915 & 0.878 & 0.988 & 0.972 & \textbf{0.992}\\
    \midrule
    Overall     & 0.697 & 0.814 & 0.730 & 0.839 & \textbf{0.854}\\
    \bottomrule
    \end{tabular}
    \end{center}
    \vspace{-10pt}
\end{table}

\begin{wrapfigure}{r}{0.25\textwidth}
\vspace{-30pt}
  \begin{center}   
  \includegraphics[width=0.25\textwidth]{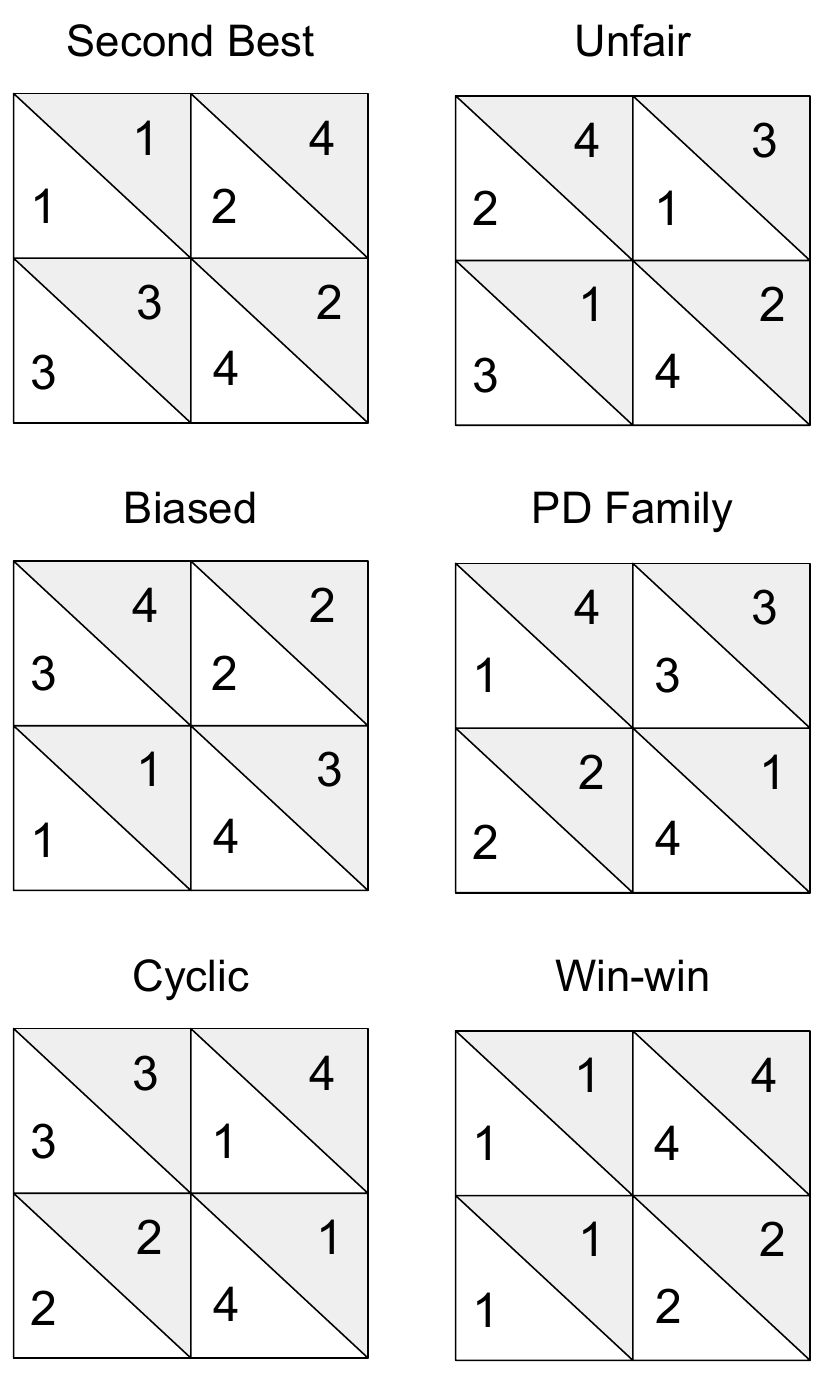}
  \end{center}
  \vspace{-10pt}
  \caption{Canonical forms of payoff matrices for each game family.}
\vspace{-20pt}
\label{fig:canonical_forms}
\end{wrapfigure}

We start out our experiments by letting the three LLMs play games from different families with each other. We focus on all known types of $2\times2$-games from the families of win-win, biased, second-best, cyclic, and unfair games as well as all games from the Prisoner's Dilemma family \cite{owen2013game, robinson2005topology}. We show example payoff matrices for each type of game in Figure \ref{fig:canonical_forms}.

We let all LLMs play with every other LLM, including themselves, for all games repeatedly over 10 rounds and with all LLMs as either Player 1 or Player 2. This leads to 1224 games in total: 324 win-win, 63 Prisoner's Dilemma, 171 unfair, 162 cyclic, 396 biased, and 108 second-best games.  Win-win games result in mutually beneficial outcomes for both players; Prisoner's Dilemma involves a conflict between individual and collective actions; unfair games have skewed outcomes favoring one player; cyclic games feature outcomes where preferences rotate; biased games have inherent advantages for one player; and second-best games involve suboptimal outcomes where no player achieves their ideal result. The sample size for each game family differs due to the specific characteristics and properties that define each family. Some families have more members due to a wider range of configurations that fit their criteria, while others have fewer games because their structural requirements are more restrictive. For example, Prisoner's Dilemma family is constrained by a structure where both players have a dominant strategy to defect, leading to a suboptimal equilibrium. On the other hand, win-win games can have multiple equilibria which provides more flexibility.

To analyze the different LLMs' performance, we calculated, for each game, their achieved score divided by the total score that could have been achieved under ideal conditions, i.e. if both players had played such that the player we are analyzing would have gained the maximum possible outcomes on every round. The results of this simulation are shown across all game types in Table~\ref{tab:all_games_results}. We can see that all models perform reasonably well. Moreover, we observe that larger LLMs generally outperform smaller LLMs. In particicular, GPT-4 performs best overall, outperforming Claude 2 ($t(287)=3.34$, $p<.001$, Cohen's $d=0.20$, 95\%-CIs=$[0.08,0.31]$, $BF=14.8$), davinci-003 ($t(287)=6.29$, $p<.001$, $d=0.37$, 95\%-CIs=$[0.25,0.49]$, $BF>100$), davinici-002 ($t(287)=8.45$, $p<.001$, $d=0.70$, 95\%-CIs=$[0.52,0.89]$, $BF>100$), and Llama 2 ($t(287)=7.27$, $p<.001$, $d=0.43$, 95\%-CIs=$[0.31,0.43]$, $BF>100$). 

We can use these results to take a glimpse at the different LLM's strengths. That LLMs are generally performing best in win-win games is not surprising, given that there is always an obvious best choice in such games. What is, however, surprising is that they also perform well in the Prisoner's Dilemma family of games, which is known to be challenging for human players \cite{jones2008smarter}. We can also use these results to look at the different LLM's weaknesses. Seemingly, all of the LLMs perform worse in situations in which what is the best choice is not aligned with their own preferences. Because humans commonly solve such games via the formation of conventions, we will look at a canonical game of convention formation, the Battle of the Sexes, in more detail below.

\subsection*{Cooperation and coordination games}

In this section, we analyze the interesting edge cases where the LLMs performed relatively well and poorly in the previous section. To do so, we take a detailed look at LLMs' behaviour in the canonical Prisoner's Dilemma and the Battle of the Sexes.

\subsubsection*{Prisoner's Dilemma} 

We have seen that LLMs perform well in games that contain elements of competition and defection. In these games, a player can cooperate with or betray their partner. When played over multiple interactions, these games are an ideal test bed to assess how LLMs retaliate after bad interactions. 

In the canonical Prisoner's Dilemma, two agents can choose to work together, i.e. cooperate, for average mutual benefit, or betray each other, i.e. defect, for their own benefit and safety. In our payoff matrix, we adhere to the general condition of a Prisoner’s Dilemma game in which the payoff relationships dictate that mutual cooperation is greater than mutual defection whereas defection remains the dominant strategy for both players:

\vspace{-10pt}
\begin{equation}
\begin{matrix}
& Cooperate & Defect \\
Cooperate & (8, 8) & (0, 10) \\
Defect & (10, 0) & (5, 5)
\end{matrix}
\label{eq:payoff_pd}
\end{equation}

Crucially, the set-up of the game is such that a rationally acting agent would always prefer to defect in the single-shot version of the game as well as in our case of finitely iterated games with knowledge of the number of trials, despite the promise of theoretically joint higher payoffs when cooperating. This is because Player 1 always runs the risk that Player 2 defects, leading to catastrophic losses for Player 1 but better outcomes for Player 2. When the game is played infinitely, however, or with an unknown number of trials, agents can theoretically profit by employing more dynamic, semi-cooperative strategies \cite{axelrod1981evolution}.

As before, we let GPT-4, text-davinci-003, text-davinci-002, Claude 2 and Llama 2 play against each other. Additionally, we introduce three simplistic strategies. Two of these strategies are simple singleton players, who either always cooperate or defect. Finally, we also introduce an agent who defects in the first round but cooperates in all of the following rounds. We introduced this agent to assess if the different LLMs would start cooperating with this agent again, signaling the potential of building trust.

\begin{figure}[t]
    \begin{center}
    \includegraphics[width=\textwidth]{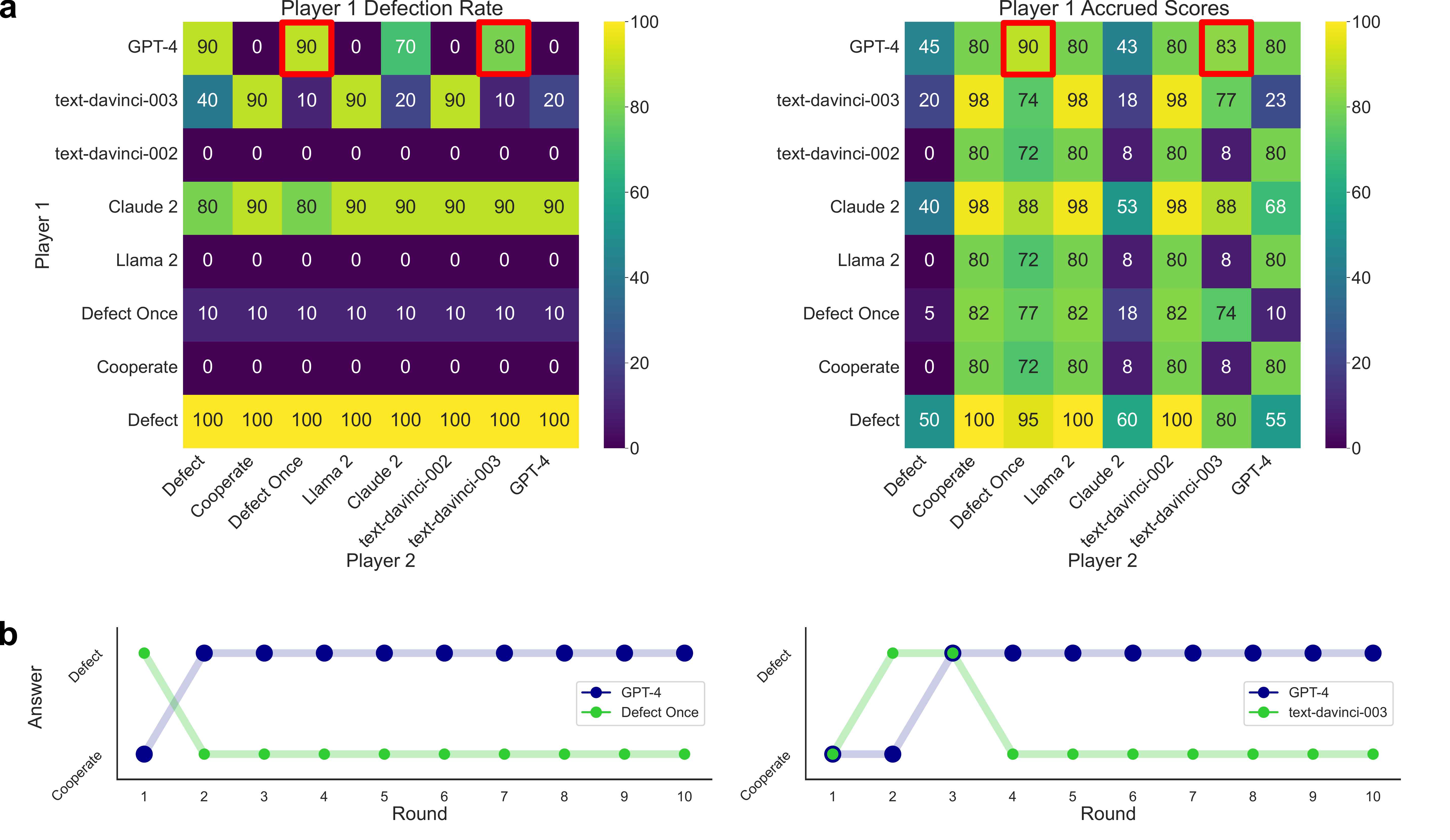}
    \end{center}
    \vspace{-10pt}
    \caption{Overview of the Prisoner's Dilemma \textbf{A:} Heatmaps showing the Player 1 defection rate in each combination of players and the scores accrued by Player 1 in each game. \textbf{B:} Example gameplays between GPT-4 and an agent that defects once and then cooperates, and between GPT-4 and text-davinci-003. These games are also highlighted in \textcolor{red}{\textbf{red}} in the heatmaps.}
    \label{fig:pd_overview}
    \vspace{-10pt}
\end{figure}

Figure \ref{fig:pd_overview} shows the results of all pairwise interactions. GPT-4 plays generally better than all other agents ($t(153.4)=3.91$, $p<.001$, $d=0.33$, 95\%-CIs=$[0.10,0.55]$, $BF=7.1$). Crucially, GPT-4 never cooperates again when playing with an agent that defects once but then cooperates on every round thereafter. Thus, GPT-4 seems to be rather unforgiving in this setup. Its strength in these families of games thus seems to generally stem from the fact that it does not cooperate with agents but mostly just chooses to defect, especially after the other agent defected once.

\paragraph{Robustness checks.} To make sure that the observed unforgivingness was not due to the particular prompt used, we run several versions of the game as robustness checks, randomising the order of the presented options, relabeling the choice options, and changing the presented utilities to be represented by either points, dollars, or coins (see Figure \ref{fig:robustness_small}). We also repeated our analysis with two different cover stories, added explicit end goals to our prompt, ran games with longer playing horizons and described numerical outcomes with text, also see Supplementary Figure 3. The results of these simulations showed that the reluctance to forgive was not due to any particular characteristics of the prompts. A crucial question was if GPT-4 did not understand that the other agent wanted to cooperate again or if it could understand the pattern but just did not act accordingly. We, therefore, run another version of the game, where we told GPT-4 explicitly that the other agent would defect once but otherwise cooperate. This resulted in GPT-4 choosing to defect throughout all rounds, thereby maximizing its own points.

\paragraph{Prompting techniques to improve observed behaviour.} One problem of these investigations in the Prisoner's Dilemma is that defecting can under specific circumstances be seen as the optimal, utility-maximizing, and equilibrium option even in a repeated version, especially if one knows that the other player will always choose to cooperate and when the number of interactions is known. Thus, we run more simulations to assess if there could be a scenario in which GPT-4 starts to forgive and cooperates again, maximizing the joint benefit instead of its own. 

We took inspiration from the literature on human forgiveness in the Prisoner's Dilemma and implemented a version of the task in the vein of \cite{fudenberg2012slow}. Specifically, \cite{fudenberg2012slow} showed that telling participants that other players sometimes make mistakes, makes people more likely to forgive and cooperate again after another player's defection (albeit in infinitely played games). Indeed, this can be favorable to them in terms of pay-offs. We observed similar behaviour in GPT-4 as it started cooperating again. 

\subsubsection*{Battle of the Sexes} 

In our large-scale analysis, we saw that the different LLMs did not perform well in games that required coordination between different players. In humans, it has frequently been found that coordination problems can be solved by the formation of conventions \cite{hawkins2016formation,young1996economics}. 

A coordination game is a type of simultaneous game in which a player will earn a higher payoff when they select the same course of action as another player. Usually, these games do not contain a pure conflict, i.e. completely opposing interests, but may contain slightly diverging rewards. Coordination games can often be solved via multiple pure strategies, or mixed, Nash equilibria in which players choose (randomly) matching strategies. Here, to probe how LLMs balance coordination and self-interest, we look at a coordination game that contains conflicting interests. 

We study a game that is archaically referred to as the ``Battle of the Sexes'', a game from the family of biased games. Assume that a couple wants to decide what to do together. Both will increase their utility by spending time together. However, while the wife might prefer to watch a football game, the husband might prefer to go to the ballet. Since the couple wants to spend time together, they will derive no utility by doing an activity separately. If they go to the ballet together, or to a football game, one person will derive some utility by being with the other person but will derive less utility from the activity itself than the other person. The corresponding payoff matrix is

\vspace{-10pt}
\begin{equation}
\begin{matrix}
& Football & Ballet \\
Football & (10, 7) & (0, 0) \\
Ballet & (0, 0) & (7, 10)
\end{matrix}
\label{eq:payoff_bos}
\end{equation}

As before, the playing agents are all three versions of GPT, Claude 2, Llama 2 as well as three more simplistic strategies. For the simplistic strategies, we implemented two agents who always choose just one option. Because LLMs most often interact with humans, we additionally implemented a strategy that mirrored a common pattern exhibited by human players in the battle of the sexes. Specifically, humans have been shown to often converge to turn-taking behaviour in the Battle of the Sexes \cite{andalman2004alternation,lau2008using,mckelvey2001playing,arifovic2018learning}; this means that players alternate between jointly picking the better option for one player and picking the option for the other player. While not a straightforward equilibrium, this behaviour has been shown to offer an efficient solution to the coordination problem involved and to lead to high joint welfare \cite{lau2008using}. 

Figure \ref{fig:bos_overview} shows the results of all interactions. As before, GPT-4 plays generally better than all other agents ($t(128.28)=2.83$, $p=.005$, $d=0.28$, 95\%-CIs=$[0.07,0.50]$, $BF=3.56$). Yet while GPT-4 plays well against other agents who choose only one option, such as an agent always choosing Football, it does not play well with agents who frequently choose their non-preferred option. For example, when playing against text-davinci-003, which tends to frequently choose its own preferred option, GPT-4 chooses its own preferred option repeatedly but also occasionally gives in and chooses the other option. Crucially, GPT-4 performs poorly when playing with an alternating pattern (where, for courtesy, we let agents start with the option that the other player preferred). This is because GPT-4 seemingly does not adjust its choices to the other player but instead keeps choosing its preferred option. GPT-4, therefore, fails to coordinate with a simple, human-like agent, an instance of a behavioural flaw.

\begin{wrapfigure}{r}{0.55\textwidth}
\vspace{-10pt}
  \begin{center}   
  \includegraphics[width=0.55\textwidth]{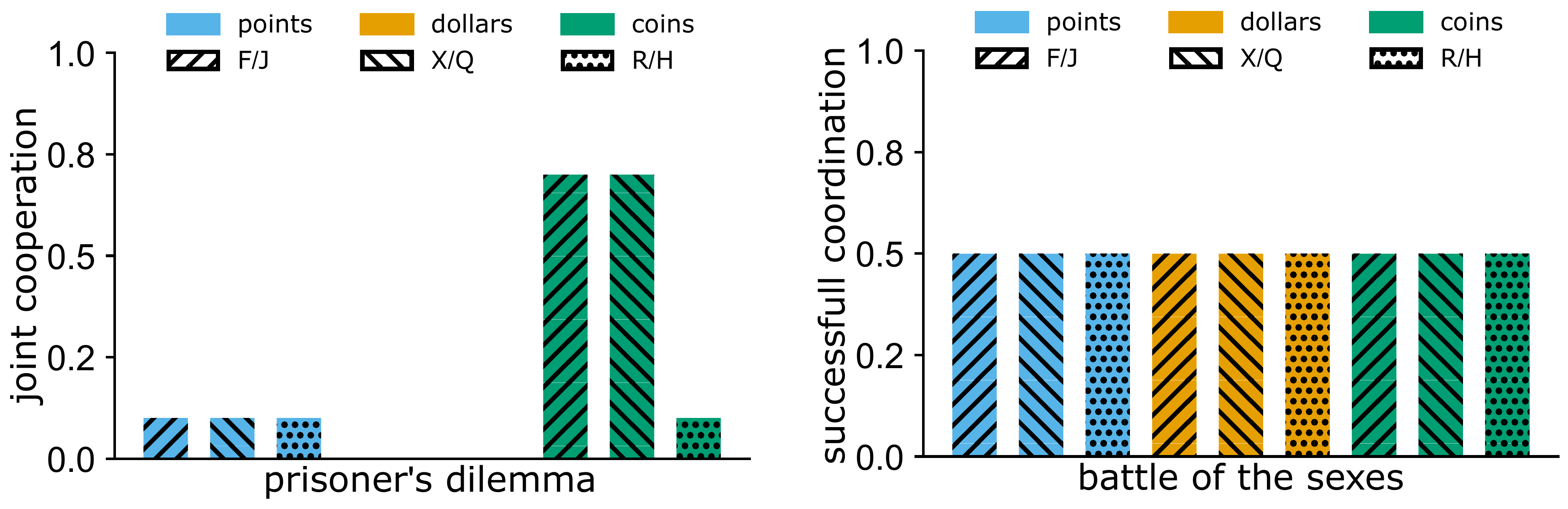}
  \end{center}
  \vspace{-10pt}
  \caption{Prompt variations. \textbf{Left:} GPT-4's performance for different prompt variations in the PD game against a false defector agent. Probability of joint cooperation $\leq0.1$ for all combinations except for two using coins as utility outcomes. \textbf{Right:} GPT-4's performance for different prompt variations in the BoS game against an alternating agent. GPT-4 always chooses its preferred option resulting in successful coordination rates of only 0.5 across all combinations. For each variation, two random letters that occur with similar frequency in English are given as the choice options.}
\label{fig:robustness_small}
\end{wrapfigure}

\paragraph{Robustness checks.} To make sure that this observed behavioural flaw was not due to the particular prompt used, we also re-run several versions of the game, where we randomize the order of the presented options, relabeled the choice options, and changed the presented utilities to be represented by either points, dollars, or coins as shown in Figure \ref{fig:robustness_small}. We also repeated our analysis with two different cover stories, in which we told GPT-4 that it was taking part in a cooking competition or working on a collaborative project keeping the underlying problem structure (payoffs and the interaction dynamics) identical (See Supplementary Figure 3). The results of these simulations showed that the inability to alternate was not due to any particular characteristics of the used prompts. To make sure that the observed behavioural flaw was not due to the particular payoff matrix used, we also re-run several versions of the game, where we modified the payoff matrix gradually from preferring Football to preferring Ballet (or, in our case, the abstract F and J). The results of these simulations showed that GPT-4 did not alternate for any of these games but simply changed its constant response to the option that it preferred for any particular game. Thus, the inability to alternate was not due to the particular payoff matrix we used (see Supplementary A.5).

\begin{figure}[t]
    \centering   \includegraphics[width=\textwidth]{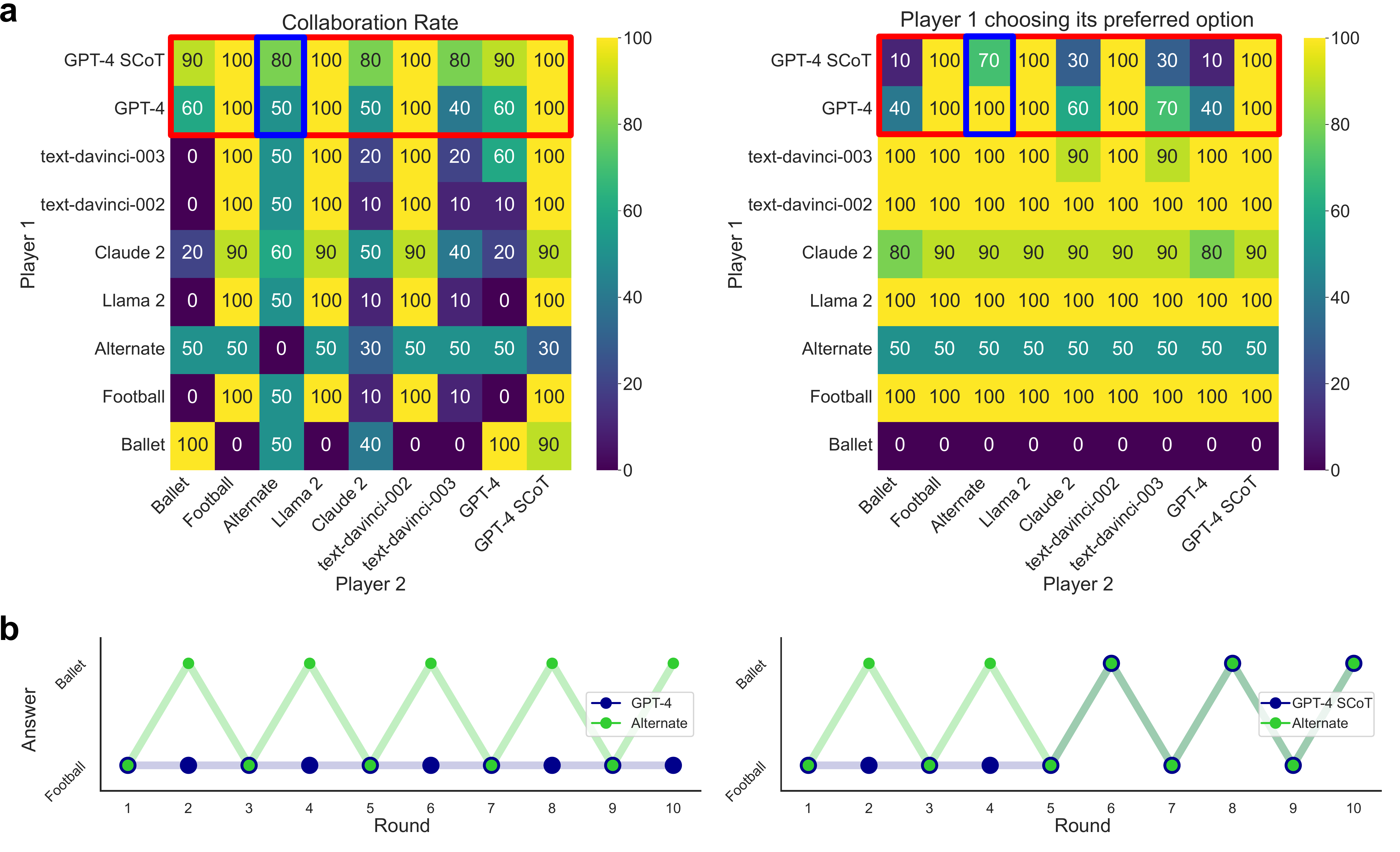}
    \vspace{-20pt}
    \caption{Overview of the Battle of the Sexes. \textbf{A:} Heatmaps showing rates of successful collaboration between the two players and the rates of Player 1 choosing its preferred option Football. GPT-4 SCoT and GPT-4 performance comparisons are highlighted in \textcolor{red}{\textbf{red}}. \textbf{B:} Gameplay between GPT-4 and an agent that alternates between the two options (\textbf{left}) and gameplay between GPT-4 and GPT-4 SCoT which represents a GPT-4 model prompted using the social chain-of-thought method to first predict the opponent’s move before making its own move by reasoning about its prediction (\textbf{right}). Both games are also highlighted in \textcolor{blue}{\textbf{blue}} in the heatmaps.}
    \label{fig:bos_overview}
    \vspace{-10pt}
\end{figure}

\begin{wrapfigure}{r}{0.4\textwidth}
\vspace{-20pt}
  \begin{center}   \includegraphics[width=0.4\textwidth]{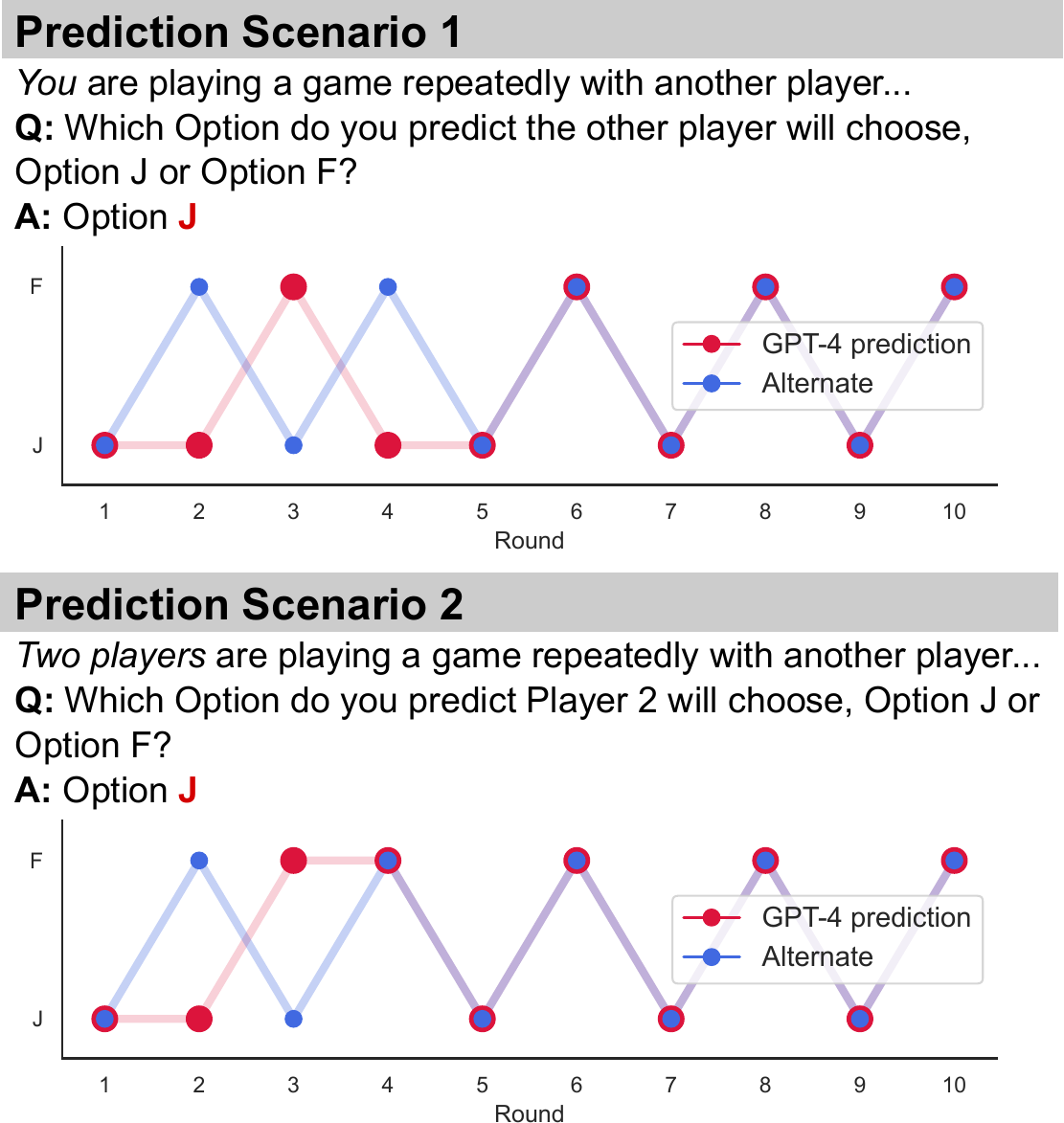}
  \end{center}
  \vspace{-10pt}
  \caption{Prediction scenarios in the Battle of the Sexes. \textbf{Top:} GPT-4 is a player of the game and predicts the other player's move. \textbf{Bottom:} GPT-4 is a mere observer of a game between Player 1 and Player 2 and predicts Player 2's move.}
\label{fig:variants_of_bos}
\end{wrapfigure}

\paragraph{Prediction scenarios.} Despite these robustness checks, another crucial question remains: Does GPT-4 simply not understand the alternating pattern or can it understand the pattern but is unable to act accordingly? To answer this question, we run two additional simulations. In the first simulation, GPT-4 was again framed as a player in the game itself. However, we now additionally ask it to predict the other player's next move according to previous rounds. In this simulation, GPT-4 started predicting the alternating pattern correctly from round 5 onward (we show this in Figure \ref{fig:variants_of_bos}).

In the second simulation, instead of having GPT-4 be framed as a player itself, we simply prompted it with a game between two ('external') players and asked it to predict one player's next move according to the previous rounds. For the shown history, we used the interaction between GPT-4 and the alternating strategy. In this simulation, GPT-4 started predicting the alternating pattern correctly even earlier, from round 3 onward. Thus, GPT-4 seemingly \emph{could} predict the alternating patterns but instead just did not act in accordance with the resulting convention. Similar divergences in abilities between social and non-social representations of the same situation have been observed in autistic children \cite{swettenham1996s}.

\paragraph{Social chain-of-thought (SCoT) prompting.} Finally, we wanted to see if GPT-4's ability to predict the other player's choices could be used to improve its own actions. This idea is closely related to how people's reasoning in repeated games and tasks about other agents' beliefs can be improved \cite{westby2014developmental}. For example, computer-aided simulations to improve the social reasoning abilities of autistic children normally include questions to imagine different actions and outcomes \cite{begeer2011theory}. This has been successfully used to improve people's decision-making more generally. It is also in line with the general finding that chain-of-thought prompting improves LLM's performance, even in tasks measuring theory of mind \cite{moghaddam2023boosting}.Thus, we implemented a version of this reasoning through actions by asking LLMs to imagine the possible actions and their outcomes before making a decision. We termed this approach social chain-of-thought prompting.
Applying this method improved GPT-4's behaviour and it started to alternate from round 5 onward (see Figure \ref{fig:bos_overview}).
    
\subsection*{Human experiments}

Given the behavioural signatures observed in GPT-4's responses in the different games, we were interested in how actual human subjects would behave when playing with such agents. To test this, we conducted an experiment in which 195 participants played both the ``Battle of the Sexes'' and the ``Prisoner's Dilemma'' against LLMs. Because the social chain-of-thought prompting turned out to be a most reliable modification of LLMs' behaviour, we only applied this prompting method in our behavioural experiments with humans.

\begin{figure}[t]
    \begin{center}
    \includegraphics[width=\textwidth]{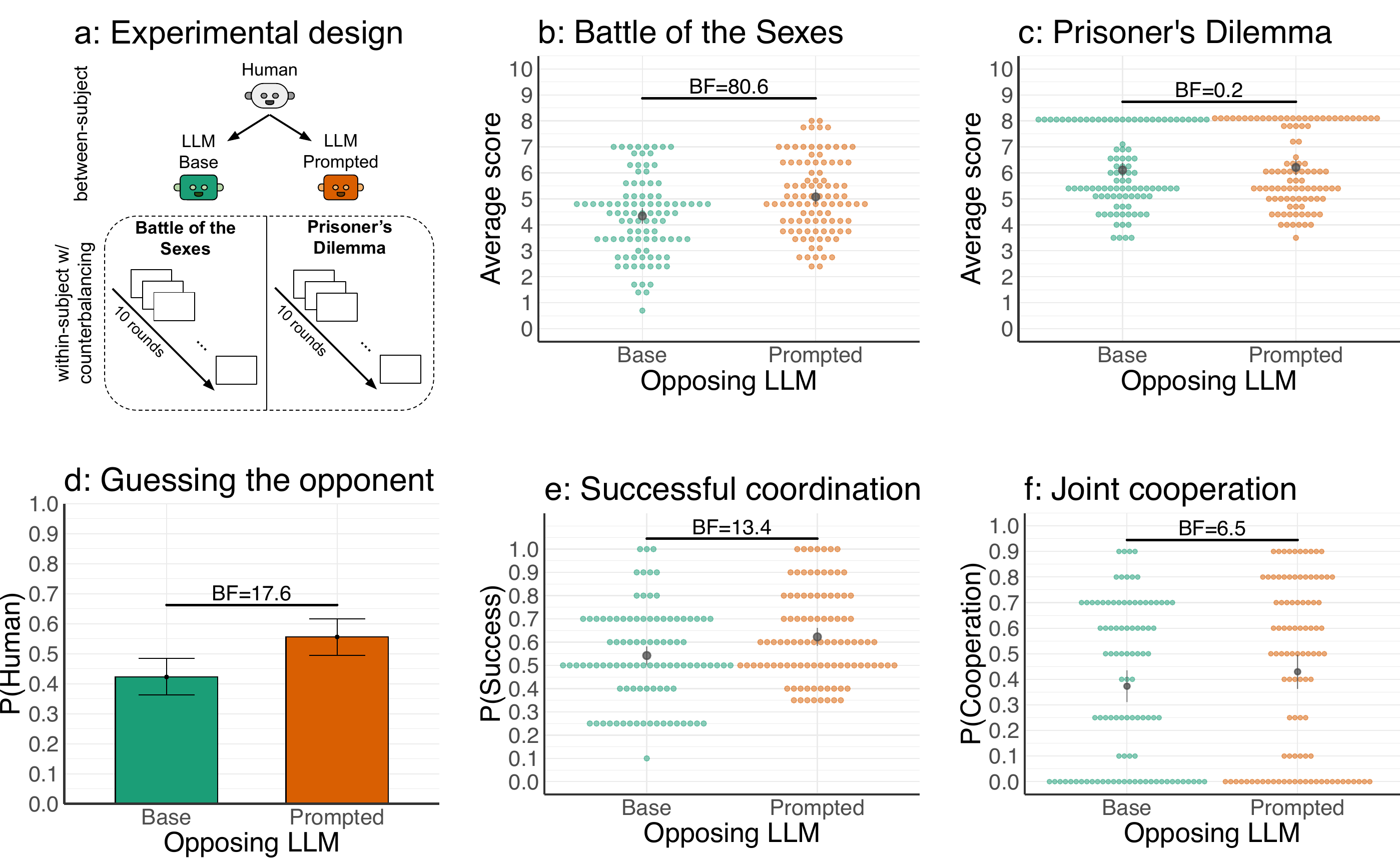}
    \end{center}
    \vspace{-10pt}
    \caption{Human experiments. \textbf{a:} Design of human experiments ($N=195$, 89 females, mean age=26.72, $SD=4.19$). Each participant gets randomly assigned either base or SCoT prompted version of the LLM at the start and plays both games repeatedly for 10 rounds against this agent. \textbf{b:} Results of the ``Battle of the Sexes'' game showing participants' average scores by condition (mixed-effects regression results: $\beta=0.74$, $t(193)=3.49$, $p<.001$, 95\%-CIs= $[0.32, 1.15]$, BF=$80.6$). \textbf{c:} Results of the ``Prisoner's Dilemma'' game showing participants' average scores by condition ($\beta=0.10$, $t(193)=0.47$, $p=0.64$, 95\%-CIs= $[-0.31, 0.51]$, BF=$0.2$). \textbf{d:} Average proportion of participants guessing that they have played against another human by condition. Error bars represent the 95\%-CIs of the mean ($\beta=0.54$, $z=8.31$, $p<.001$, 95\%-CIs= $[0.05, 0.42]$, BF=$17.6$). \textbf{e:} Participants' successful coordination rates by condition in the ``Battle of the Sexes'' game ($\beta=0.33$, $z=3.59$, $p<.001$, 95\%-CIs= $[0.15, 0.51]$, BF=$13.4$). \textbf{f:} Participants' mutual cooperation rates by condition in the ``Prisoner's Dilemma'' game ($\beta=0.24$, $z=2.54$, $p=0.01$, 95\%-CIs= $[0.05, 0.42]$, BF=$6.5$). }
    \label{fig:human}
    \vspace{-10pt}
\end{figure}

Participants were told that they would play either against a human player or an artificial agents for 10 repeated rounds for each game and, after each game, had to guess whether they had played against a human or not. Which game they played first was assigned randomly. While all subjects, in fact, only played against LLMs, one group played against the base version of GPT-4, while another group played against a version of GPT-4 that first predicted the other agent's move and the acted accordingly, i.e. social chain-of-thought prompting. Importantly, each participant played only two games, and the prompting was reset between games to ensure any change in LLM behavior was not influenced by prior interactions within the experiment. If assigned to the base version initially, participants played both games with this model, and likewise for the socially prompted version. An overview of the experimental design can be seen in Figure \ref{fig:human}a. Participants were recruited from Prolific and debriefed fully after the experiment. We were interested in how people played against LLMs in general as well as if GPT-4's behaviour could be improved via social chain-of-thought prompting. Finally, we also asked participants whether they thought they had played with another human or an artificial agent after each game. 

While participants' average score was significantly higher for the SCoT-prompted condition compared to the condition without further prompting (i.e. base) in the Battle of the Sexes (mixed-effects regression results: $\beta=0.74$, $t(193)=3.49$, $p<.001$, 95\%-CIs= $[0.32, 1.15]$, BF=$80.6$), no such difference was observed in the Prisoner's Dilemma ($\beta=0.10$, $t(193)=0.47$, $p=0.64$, 95\%-CIs= $[-0.31, 0.51]$, BF=$0.2$). Looking at the behaviour of both players, we found that SCoT-prompting increased successful coordination (i.e. both players picking the same option) in the Battle of the Sexes ($\beta=0.33$, $z=3.59$, $p<.001$, 95\%-CIs= $[0.15, 0.51]$, BF=$13.4$), while it also slightly increased joint cooperation (i.e. both players cooperating) in the Prisoner's Dilemma ($\beta=0.24$, $z=2.54$, $p=0.01$, 95\%-CIs= $[0.05, 0.42]$, BF=$6.5$). In general, participants were more likely to think that the prompted model was another human player as compared to the unprompted base GPT-4 model ($\beta=0.54$, $z=8.31$, $p<.001$, 95\%-CIs= $[0.05, 0.42]$, BF=$17.6$). Additional analysis on participants' temporal behaviour in both games can be found in the SI.

In summary, SCoT prompting can increase GPT-4's coordination and cooperation behaviour without changing scores in scenarios where self-interest is important for good behaviour, i.e. the Prisoner's Dilemma, but leading to increased performance in coordination problems, i.e. the Battle of the Sexes. 

\section*{Discussion}

LLMs are one of the most quickly adopted technologies ever, interacting with millions of consumers within weeks \cite{bommasani2021opportunities}. Understanding in a more principled manner how these systems interact with us, and with each other, is thus of urgent concern. Here, our proposal is simple: Just like behavioural game theorists use tightly controlled and theoretically well-understood games to understand human interactions, we use these games to study the interactions of LLMs.

We thereby understand our work as both a proof of concept of the utility of this approach and an examination of the individual failures and successes of socially interacting LLMs. Our large-scale analysis of all $2\times2$-games highlights that the most recent LLMs indeed are able to perform well on a wide range of game-theoretic tasks as measured by their own individual reward, particularly when they do not have to explicitly coordinate with others. This adds to a wide-ranging literature showcasing emergent phenomena in LLMs \cite{brown2020language, wei2022emergent, webb2022emergent, chen2021evaluating, bubeck2023sparks}. However, we also show that LLMs behaviour is suboptimal in coordination games, even when faced with simple strategies.

To tease apart the behavioural signatures of these LLMs, we zoomed in on two of the most canonical games in game theory: the Prisoner's Dilemma and the Battle of the Sexes. In the Prisoner's Dilemma, we show that GPT-4 mostly plays unforgivingly. Starting with full cooperation, it permanently shifts to defection after a single negative interaction with the other agent, even if the other agent later cooperates. While noting that GPT-4's continual defection is indeed the equilibrium policy in this finitely played game, such behaviour comes at the cost of the two agents' joint payoff. We see a similar tendency in GPT-4's behaviour in the Battle of the Sexes, where it has a strong tendency to stubbornly stick with its own preferred alternative. In contrast to the Prisoner's Dilemma, this behaviour is suboptimal, even on the individual level. 

Current generations of LLMs are generally assumed, and trained, to be benevolent assistants to humans \cite{ouyang2022training}. Despite many successes in this direction, the fact that we here show how they play iterated games in such a selfish, and uncoordinated manner sheds light on the fact that there is still significant ground to cover for LLMs to become truly social and well-aligned machines \cite{wolf2023fundamental}. Their lack of appropriate responses vis-a-vis even simple strategies in coordination games also speaks to the recent debate around theory of mind in LLMs \cite{ullman2023large, le2019revisiting, kosinski2023theory} by highlighting a potential failure mode.

Our extensive robustness checks demonstrate how these behavioural signatures are not functions of individual prompts but reflect broader patterns of LLM behaviour. Our intervention pointing out the fallibility of the playing partner -- which leads to increased cooperation -- adds to a literature that points to the malleability of LLM social behaviour in tasks to prompts  \cite{horton2023large, aher2022using}. This is important as we try to understand what makes LLMs better, and more pleasant, interactive partners. Further experiments on GPT-4's final round behaviour have shown that it did not adjust its behavior in the last round of games or when faced with varying probabilities of continuation, unlike human players who often increase cooperation when future interactions are likely \cite{dalbo2011evolution,nowak2005evolution}. This suggests that GPT-4 may lack mechanisms for backward induction and long-term strategic planning, primarily focusing on immediate context due to its training on next-token prediction \cite{radford2019language}. Consequently, GPT-4 tends to default to defection in uncertain situations, contrasting with human tendencies to anticipate and adjust based on future outcomes \cite{axelrod1981evolution,nowak2006five}. 

We additionally observed that prompting GPT-4 to make predictions about the other player before making its own decisions can alleviate behavioural flaws and the oversight of even simple strategies. This represents a more explicit way to force an LLM to engage in theory of mind and shares much overlap with non-social chain-of-thought reasoning \cite{wei2022chain, moghaddam2023boosting}. Just like chain-of-thought prompting is now implemented as a default in some LLMs to improve (non-social) reasoning performance, our work suggests implementing a similar social cognition prompt to improve human-LLM interaction.

In our exploration of a behavioural game theory of machines, we acknowledge several limitations. First, despite covering many families of games, our investigation is constrained to simple $2\times2$ games. However, we note that our analysis significantly goes beyond current investigations that have often investigated only one game, and done so using single-shot rather than iterated instances of these games. For example, our iterated approach shares more overlap with the more iterated nature of human-LLM conversations. We also note that we mainly study finite games where agents share knowledge about the duration of the interaction. This is in contrast to so-called indefinite games that have either unknown, probabilistic or no endpoints at all. In these games, both optimal prescriptions and empirical behaviour can differ significantly from the finite case, warranting further investigation.

We believe that more complicated games will shed even more light on game-theoretic machine behaviour in the future. For example, games with more continuous choices like the trust game \cite{engle2004evolution} might elucidate how LLMs dynamically develop (mis-)trust. Games with more than two agents, like public goods or tragedy of the commons type games \cite{rankin2007tragedy} could probe how 'societies' of LLMs behave, and how LLMs cooperate or exploit each other.

Given the social nature of the tasks studied here, further empirical work is needed to fully understand human-LLM interactions across all paradigms. In our study, we conducted human experiments in two of the games, specifically, the Battle of the Sexes and the Prisoner's Dilemma, and attempted to probe human-like behaviors such as turn-taking in battle Battle of the Sexes or prompting for forgiveness in the Prisoner's Dilemma. However, these empirical investigations were limited to these two games. By extending human studies to the remaining games, additional dynamics may emerge. Furthermore, asking LLMs to self-report their strategies in these games and correlating these explanations with their actions could provide valuable insights into their actual decision-making processes. 

Our results highlight the broader importance of a behavioural science for machines \cite{rahwan2022machine,schulz2020computational, binz2023using, coda2023inducing}. We believe that these methods will continue to be useful for elucidating the many facets of LLM cognition, particularly as these models become more complex, multi-modal, and embedded in physical systems.

\subsection*{Related work}

As algorithms become increasingly more able and their decision making processes impenetrable, the behavioural sciences offer new tools to make inferences just from behavioural observations \cite{rahwan2022machine,schulz2020computational}. behavioural tasks have, therefore, been used in several benchmarks \cite{bommasani2021opportunities,kojima2022large}.

Whether and how algorithms can make inferences about other agents, machines and otherwise, is one stream of research that borrows heavily from the behavioural sciences \cite{rabinowitz2018machine, cuzzolin2020knowing,alon2022dis}. Of particular interest to the social interactions most LLMs are embedded in is an ability to reason about the beliefs, desires, and intentions of other agents, or a so-called theory of mind (ToM) \cite{frith2005theory}. Theory of mind underlies a wide range of interactive phenomena, from benevolent teaching \cite{velez2021learning} to malevolent deception \cite{lissek2008cooperation, alon2022dis}, and is thought to be the key to many social phenomena in human interactions \cite{hula2015monte, ho2022planning}. 

Whether LLMs possess a theory of mind has been debated. For example, it has been argued that GPT-3.5 performs well on a number of canonical ToM tasks \cite{kosinski2023theory}. Others have contested this view, arguing that such good performance is merely a function of the specific prompts \cite{ullman2023large,le2019revisiting}. Yet other research has shown that chain-of-thought reasoning significantly improves LLMs' ToM ability \cite{moghaddam2023boosting}. Moreover, the currently largest LLM, GPT-4, manages to perform well in ToM tasks, including in the variants in which GPT-3.5 previously struggled \cite{bubeck2023sparks}. Thus, GPT-4's behaviour will be of particular interest in our experiments.

Games taken from game theory present an ideal testbed to investigate interactive behaviour in a controlled environment\cite{han2021or} and LLM's behaviour has been probed in such tasks \cite{chan2023towards}. For example, \cite{horton2023large} let GPT-3 participate in the dictator game, and \cite{aher2022using} used the same approach for the ultimatum game. Both show how the models' behaviour is malleable to different prompts, for example making them more or less self-interested. However, all these games rely on single-shot interactions over fewer games and do not use iterated games.

Our study builds upon recent advancements in the field, which have shifted the focus from solely assessing the performance of LLMs to comparing them with human behaviours. Previous research efforts have explored various approaches to analyze LLMs, such as employing cognitive psychology tools \cite{binz2023using,dasgupta2022language} and even adopting a computational psychiatry perspective \cite{coda2023inducing}.

Finally, the theory behind interacting agents is important for many machine learning applications in general \cite{crandall2011learning}, and in particular, in adversarial settings \cite{goodfellow2020generative}, where one agent tries to trick the other agent into thinking that a generated output is good. Understanding prosocial dynamics in multiagent systems\cite{santos2024prosocial}, and fostering cooperation in them\cite{guo2023facilitating} is essential for developing robust and trustworthy AI systems that can navigate complex social environments\cite{powers2023stuff}.

\section*{Methods}

\label{general_approach}

To investigate how human subjects would behave when playing with LLM agents, we studied their interactions in two of the games we used; Prisoner's Dilemma and the Battle of the Sexes. We also investigated if participants could detect and behave differently when playing against different agents. Participants (N=195, 89 females, mean age=26.72, SD=4.19) were recruited through Prolific\cite{palan2018prolific}, an online platform that allows researchers to access a diverse and reliable pool of participants. No statistical methods were used to pre-determine sample sizes but our sample sizes are similar to those reported in previous publications\cite{normann2012impact, charness2002understanding, wong2005dynamic}. The participants were required to be fluent speakers of English with minimum approval rates of .95 and 1, and a minimal number of previous submissions of 10 that have not participated in our experiment before. All participants provided informed consent prior to inclusion in the study. Experiments were performed in accordance with the relevant guidelines and regulations approved by the ethics committee of the University of T\"ubingen (protocol nr. 701/2020BO). Participants received a £3 base payment plus a bonus of up to £2 depending on performance (1 Cent for each point received during the games) for their participation. The average compensation was £11.41 per hour. Participants were fully debriefed after the experiment. Data of 21 players who failed to make a round's choice between the 2 options within a given time frame (20 seconds) were excluded. 

In the sections that follow, we first detail the experimental setup for LLM-LLM interactions, which serves as a comparative baseline for our study. We then present details from the human participant study outlined above.

\subsection*{LLM-LLM Interactions}

We study LLMs' behaviour in finitely repeated games with full information taken from the economics literature. We focus on two-player games with discrete choices between two options to simplify the analyses of emergent behaviours. We let two LLMs interact via prompt-chaining, i.e. all integration of evidence and learning about past interactions happens as in-context learning \cite{brown2020language, liu2023pre}. The games are submitted to LLMs as prompts in which the respective game, including the choice options, is described. At the same time, we submit the same game as a prompt to another LLM. We obtain generated tokens $\boldsymbol{t}$ from both LLMs by sampling from
\begin{equation}
    p_{\text{LLM}}(\boldsymbol{t}|\boldsymbol{c}^{(p)}) = \prod_{k=1}^K p_{\text{LLM}}(t_k|c_1^{(p)}, \dots, c_n^{(p)}, t_1,...,t_{k-1})
\end{equation}

After feeding the prompt to the LLM, our methodology is as follows. The LLM prediction of the first token following the context is $d = p_{\text{LLM}}(t_1|\boldsymbol{c}^{(p)})$ and the $N$ tokens for the possible answers of the multiple choice question are $o = \{o_i\}_{i=1}^N$ which in this case are J and F. The predicted option is then given by
\begin{equation}
    \hat{o} = \arg \max(\hat{c}_i), \text{ with } \hat{c}_i = d[c_i], i=1...N
\end{equation}

which are the predicted probabilities of the language model. Once both LLMs have made their choices, which we track as a completion of the given text, we update the prompts with the history of past interactions as concatenated text and then submit the new prompt to both models for the next round. These interactions continue for 10 rounds in total for every game. In a single round $\pi_i(x_{1}, x_{2})$ is the payoff for Player 1 when $x_{1}$ and $x_{2}$ are the strategies chosen by both players. In repeated games, the payoffs are often considered as discounted sums of the payoffs in each game stage, using a discount factor $\delta$. If the game is repeated $n$ times, the payoff $U_i$ for player $i$ is
\begin{equation}
    U_i = \pi_i(x_{10}, x_{20}) + \delta \cdot \pi_i(x_{11}, x_{21}) + \delta^2 \cdot \pi_i(x_{12}, x_{22}) + \ldots + \delta^{n-1} \cdot \pi_i(x_{1(n-1)}, x_{2(n-1)})
\end{equation}
Each term represents the discounted payoff at each stage of the repeated game, from the first game $(t=0)$ to the $n^{th}$ game $(t=n-1)$. In our experiments we keep $\delta=1$. To avoid influences of the particular framing of the scenarios, we only provide barebones descriptions of the payoff matrices (see example in Figure \ref{fig:overview}). To avoid contamination through particular choice names or the used framing, we use the neutral options `F' and `J' throughout \cite{binz2023using}. 

\paragraph{Games considered.} We first investigate 144 different $2\times2$-games where each player has two options, and their individual reward is a function of their joint decision. These games can be categorized into six distinct families\textit{—Win-Win, Prisoner’s Dilemma Family, Unfair, Cyclic, Biased, and Second-Best—}each with unique characteristics and outcomes. A win-win game is a special case of a non-zero-sum game that produces a mutually beneficial outcome for both players provided that they choose their corresponding best option. They encourage cooperation, leading to outcomes where both parties benefit. Briefly, in games from the Prisoner's Dilemma family, two agents can choose to work together, i.e. cooperate, for average mutual benefit, or betray each other, i.e. defect, for their own benefit. The typical outcome is a Nash Equilibrium that is suboptimal for both players compared to a possible Pareto-superior outcome. In an unfair game, one player can always win when playing properly, leading to highly unequal outcomes. Cyclic games are characterized by the absence of dominant strategies and equilibria. In these games, players can cycle through patterns of choices without settling into a stable outcome. Biased games are games where agents get higher points for choosing the same option but where the preferred option differs between the two players. One form of a biased game is the Battle of the Sexes, where players need to coordinate to choose the same option. Finally, second-best games are games where both agents fare better if they jointly choose the option that has the second-best utility. In many of these games, strategic swaps in payoffs can alter the game dynamics, potentially converting them into different types of games. For two additional games, Prisoner's Dilemma and Battle of the Sexes, we also let LLMs play against simple, hand-coded strategies to understand their behaviour in more detail.

\paragraph{Large Language Models considered.} In this work, we evaluate five LLMs. For all of our tasks, we used the public OpenAI API with the GPT-4, \texttt{text-davinci-003} and \texttt{text-davinci-002} models which are available via the completions endpoint, Meta AI's Llama 2 70B chat model which has 70 billion parameters and is optimized for dialogue use cases, and the Anthropic API model Claude 2 to run our simulations. Experiments with other popular open source models MosaicPretrainedTransformer (MPT), Falcon and different versions of Llama 2 (namely \texttt{MPT-7B, MPT-30B, Falcon-7b, Falcon-40b, Llama 2 7B, Llama 2 13B}) have revealed that these models did not perform well at the given tasks, choosing the first presented option more than 95\% of the time independent of which option this is. Therefore, we chose not to include them in our main experiments. For all models, we set the temperature parameters to 0 and only ask for one token answer to indicate which option an agent would like to choose. All other parameters are kept as default values.

\paragraph{Playing 6 families of $2\times2$-games task design.}

While $2\times2$-games games can appear simple, they present some of the most powerful ways to probe diverse sets of interactions, from pure competition to mixed motives and cooperation - which can further be classified into canonical subfamilies outlined elegantly by \cite{robinson2005topology}. Here, to cover the wide range of possible interactions, we study the behaviours of GPT-4, text-davinci-003, text-davinci-002, Claude 2 and Llama 2 across these canonical families. We let all five engines play all variants of games from within the six families. 

\paragraph{Cooperation and coordination task design.}

We then analyze two games, Prisoner's Dilemma and Battle of the Sexes, in more detail because they represent interesting edge cases where the LLMs performed exceptionally well, and relatively poorly. We particularly hone in on GPT-4's behaviour because of recent debates around its ability for theory of mind, that is whether it is able to hold beliefs about other agents' intentions and goals, a crucial ability to successfully navigate repeated interactions \cite{bubeck2023sparks, kosinski2023theory}. For the two additional games, we also let LLMs play against simple, hand-coded strategies to further understand their behaviour. These simple strategies are designed to assess how LLMs behave when playing with more human-like players.

\paragraph{Statistical tests.}
All reported tests are two-sided. We also report Bayes Factors quantifying the likelihood of the data under $H_A$ relative to the likelihood of the data under $H_0$. We calculate the default two-sided Bayesian \textit{t}-test using a Jeffreys-Zellner-Siow prior with its scale set to $\sqrt{2}/2$, following \cite{rouder2009bayesian}. For parametric tests, the data distribution was assumed to be normal but this was not formally tested. We report effect sizes as either Cohen's d or standardized regression estimates, including their 95\%-Confidence Intervals. 

\subsection*{Human-LLM Interactions}
Following sections provide additional details on the design and conduct of the human participant study; including compensation, demographics, prompting and the cover stories.

\paragraph{Design.} Experiments were presented to participants using a combination of HTML, JavaScript, and CSS with custom code. After a presentation of the instructions including screenshots from the actual game-play, participants were required to complete a comprehension questionnaire. Only upon responding correctly to all questions, they could proceed to the main part
of the experiment. Participants played both the Prisoner’s Dilemma and the Battle of the Sexes, with the order counter-balanced between subjects. Participants were instructed that they would play two games with 10 rounds each with different players. The participants' interface (Supplementary Figure 4) was designed to provide clear and actionable information about the current game. After each game, participants were asked to indicate if they thought they had just played with another human player or an artificial agent. 

\paragraph{Prompts and human instructions.} The cover story used for interactions with both LLMs and human participants was content-wise identical, including the rules of the game and the history of previous interactions, to ensure consistent framing across conditions (see Supplementary A.1 for the detailed prompt progression). However, the presentation was adapted to suit each audience. For human participants, visual cues and concise text were prioritized to create a more engaging experience (Supplementary Figure 4).

\paragraph{Ending and debriefing.} Participants were informed that their opponent could either be another human participant or an artificial agent. In reality, all participants were paired with either a SCoT-prompted or unprompted version of GPT-4 for the entirety of the experiment, i.e., across both games. After completing the study, participants were debriefed that the purpose of the study was to explore how to make Large Language Models (LLMs) more human-like and that, in both games, they had played against different versions of an artificial agent.

\section*{Data availability}

All participant and model simulation data from the experiments are publicly available on GitHub (\href{https://github.com/eliaka/repeatedgames}{github.com/eliaka/repeatedgames}).

\section*{Code availability}

The code underlying this study, prompt variations and model simulations are available on \href{https://github.com/eliaka/repeatedgames}{github.com/eliaka/repeatedgames}.

\section*{Acknowledgements}

This work was supported by grants from the Max Planck Society (E.A., L.S., J.C.F. and E.S.), the Volkswagen Foundation (E.S.), the German Federal Ministry of Education and Research (BMBF): Tübingen AI Center, FKZ: 01IS18039A, and the Deutsche Forschungsgemeinschaft (DFG, German Research Foundation) under Germany’s Excellence Strategy – EXC 2064/1 – grant no. 390727645 (E.A., S.J.O. and M.B). The funders had no role in study design, data collection and analysis, decision to publish or preparation of the manuscript. We thank the International Max Planck Research School for Intelligent Systems (IMPRS-IS) for supporting E.A.

\section*{Author contributions}

E.A., L.S. and E.S. conceived experiments. E.A. conducted the experiments. E.A. and E.S. analysed the results with input from L.S., J.C.F. and M.B.  E.A., L.S and E.S. wrote the manuscript with input from S.J.O. and M.B. All authors reviewed the manuscript.

\section*{Competing interests statement}
The authors declare no competing interests.


\newpage

\begin{thebibliography}{10}
\urlstyle{rm}
\expandafter\ifx\csname url\endcsname\relax
  \def\url#1{\texttt{#1}}\fi
\expandafter\ifx\csname urlprefix\endcsname\relax\def\urlprefix{URL }\fi
\expandafter\ifx\csname doiprefix\endcsname\relax\def\doiprefix{DOI: }\fi
\providecommand{\bibinfo}[2]{#2}
\providecommand{\eprint}[2][]{\url{#2}}

\bibitem{brants2007large}
\bibinfo{author}{Brants, T.}, \bibinfo{author}{Popat, A.}, \bibinfo{author}{Xu, P.}, \bibinfo{author}{Och, F.~J.} \& \bibinfo{author}{Dean, J.}
\newblock \bibinfo{title}{Large language models in machine translation}.
\newblock In \emph{\bibinfo{booktitle}{Proceedings of the 2007 Joint Conference on Empirical Methods in Natural Language Processing and Computational Natural Language Learning (EMNLP-CoNLL)}}, \bibinfo{pages}{858--867} (\bibinfo{year}{2007}).

\bibitem{devlin2018bert}
\bibinfo{author}{Devlin, J.}, \bibinfo{author}{Chang, M.-W.}, \bibinfo{author}{Lee, K.} \& \bibinfo{author}{Toutanova, K.}
\newblock \bibinfo{journal}{\bibinfo{title}{Bert: Pre-training of deep bidirectional transformers for language understanding}}.
\newblock {\emph{\JournalTitle{arXiv preprint arXiv:1810.04805}}}  (\bibinfo{year}{2018}).

\bibitem{radford2018improving}
\bibinfo{author}{Radford, A.}, \bibinfo{author}{Narasimhan, K.}, \bibinfo{author}{Salimans, T.}, \bibinfo{author}{Sutskever, I.} \emph{et~al.}
\newblock \bibinfo{journal}{\bibinfo{title}{Improving language understanding by generative pre-training}}.
\newblock {\emph{\JournalTitle{OpenAI}}}  (\bibinfo{year}{2018}).

\bibitem{brown2020language}
\bibinfo{author}{Brown, T.} \emph{et~al.}
\newblock \bibinfo{journal}{\bibinfo{title}{Language models are few-shot learners}}.
\newblock {\emph{\JournalTitle{Advances in neural information processing systems}}} \textbf{\bibinfo{volume}{33}}, \bibinfo{pages}{1877--1901} (\bibinfo{year}{2020}).

\bibitem{wei2022emergent}
\bibinfo{author}{Wei, J.} \emph{et~al.}
\newblock \bibinfo{journal}{\bibinfo{title}{Emergent abilities of large language models}}.
\newblock {\emph{\JournalTitle{arXiv preprint arXiv:2206.07682}}}  (\bibinfo{year}{2022}).

\bibitem{webb2022emergent}
\bibinfo{author}{Webb, T.}, \bibinfo{author}{Holyoak, K.~J.} \& \bibinfo{author}{Lu, H.}
\newblock \bibinfo{journal}{\bibinfo{title}{Emergent analogical reasoning in large language models}}.
\newblock {\emph{\JournalTitle{Nature Human Behaviour}}} \textbf{\bibinfo{volume}{7}}, \bibinfo{pages}{1526--1541} (\bibinfo{year}{2023}).

\bibitem{chen2021evaluating}
\bibinfo{author}{Chen, M.} \emph{et~al.}
\newblock \bibinfo{journal}{\bibinfo{title}{Evaluating large language models trained on code}}.
\newblock {\emph{\JournalTitle{arXiv preprint arXiv:2107.03374}}}  (\bibinfo{year}{2021}).

\bibitem{bubeck2023sparks}
\bibinfo{author}{Bubeck, S.} \emph{et~al.}
\newblock \bibinfo{journal}{\bibinfo{title}{Sparks of artificial general intelligence: Early experiments with gpt-4}}.
\newblock {\emph{\JournalTitle{arXiv preprint arXiv:2303.12712}}}  (\bibinfo{year}{2023}).

\bibitem{codaforno2023metaincontext}
\bibinfo{author}{Coda-Forno, J.} \emph{et~al.}
\newblock \bibinfo{journal}{\bibinfo{title}{Meta-in-context learning in large language models}}.
\newblock {\emph{\JournalTitle{Advances in Neural Information Processing Systems}}} \textbf{\bibinfo{volume}{36}}, \bibinfo{pages}{65189--65201} (\bibinfo{year}{2023}).

\bibitem{bommasani2021opportunities}
\bibinfo{author}{Bommasani, R.} \emph{et~al.}
\newblock \bibinfo{journal}{\bibinfo{title}{On the opportunities and risks of foundation models}}.
\newblock {\emph{\JournalTitle{arXiv preprint arXiv:2108.07258}}}  (\bibinfo{year}{2021}).

\bibitem{fudenberg2012slow}
\bibinfo{author}{Fudenberg, D.}, \bibinfo{author}{Rand, D.~G.} \& \bibinfo{author}{Dreber, A.}
\newblock \bibinfo{journal}{\bibinfo{title}{Slow to anger and fast to forgive: Cooperation in an uncertain world}}.
\newblock {\emph{\JournalTitle{American Economic Review}}} \textbf{\bibinfo{volume}{102}}, \bibinfo{pages}{720--749} (\bibinfo{year}{2012}).

\bibitem{mailath2004coordination}
\bibinfo{author}{Mailath, G.~J.} \& \bibinfo{author}{Morris, S.}
\newblock \bibinfo{journal}{\bibinfo{title}{Coordination failure in repeated games with almost-public monitoring}}.
\newblock {\emph{\JournalTitle{Available at SSRN 580681}}}  (\bibinfo{year}{2004}).

\bibitem{camerer2011behavioral}
\bibinfo{author}{Camerer, C. F.},
\newblock \bibinfo{book}{\bibinfo{title}{Behavioral game theory: Experiments in strategic interaction}}.
\newblock {\emph{\JournalTitle{Princeton university press}}} (\bibinfo{year}{2011}).

\bibitem{fudenberg1991game}
\bibinfo{author}{Fudenberg, D.} \& \bibinfo{author}{Tirole, J.}
\newblock \emph{\bibinfo{title}{Game theory}} (\bibinfo{publisher}{MIT press}, \bibinfo{year}{1991}).

\bibitem{VNM}
\bibinfo{author}{Von~Neumann, J.} \& \bibinfo{author}{Morgenstern, O.}
\newblock \bibinfo{title}{Theory of games and economic behavior}.
\newblock In \emph{\bibinfo{booktitle}{Theory of games and economic behavior}} (\bibinfo{publisher}{Princeton university press}, \bibinfo{year}{1944}).

\bibitem{camerer1997progress}
\bibinfo{author}{Camerer, C.~F.}
\newblock \bibinfo{journal}{\bibinfo{title}{Progress in behavioral game theory}}.
\newblock {\emph{\JournalTitle{Journal of economic perspectives}}} \textbf{\bibinfo{volume}{11}}, \bibinfo{pages}{167--188} (\bibinfo{year}{1997}).

\bibitem{henrich2001search}
\bibinfo{author}{Henrich, J.} \emph{et~al.}
\newblock \bibinfo{journal}{\bibinfo{title}{In search of homo economicus: behavioral experiments in 15 small-scale societies}}.
\newblock {\emph{\JournalTitle{American Economic Review}}} \textbf{\bibinfo{volume}{91}}, \bibinfo{pages}{73--78} (\bibinfo{year}{2001}).

\bibitem{rousseau1998not}
\bibinfo{author}{Rousseau, D.~M.}, \bibinfo{author}{Sitkin, S.~B.}, \bibinfo{author}{Burt, R.~S.} \& \bibinfo{author}{Camerer, C.}
\newblock \bibinfo{journal}{\bibinfo{title}{Not so different after all: A cross-discipline view of trust}}.
\newblock {\emph{\JournalTitle{Academy of management review}}} \textbf{\bibinfo{volume}{23}}, \bibinfo{pages}{393--404} (\bibinfo{year}{1998}).

\bibitem{johnson2022measuring}
\bibinfo{author}{Johnson, T.} \& \bibinfo{author}{Obradovich, N.}
\newblock \bibinfo{journal}{\bibinfo{title}{Measuring an artificial intelligence agent's trust in humans using machine incentives}}.
\newblock {\emph{\JournalTitle{arXiv preprint arXiv:2212.13371}}}  (\bibinfo{year}{2022}).

\bibitem{openai2023gpt4}
\bibinfo{author}{Achiam, J.} \emph{et~al.}
\newblock \bibinfo{journal}{\bibinfo{title}{Gpt-4 technical report}}.
\newblock {\emph{\JournalTitle{arXiv preprint arXiv:2303.08774}}}  (\bibinfo{year}{2023}).

\bibitem{owen2013game}
\bibinfo{author}{Owen, G.}
\newblock \emph{\bibinfo{title}{Game theory}} (\bibinfo{publisher}{Emerald Group Publishing}, \bibinfo{year}{2013}).

\bibitem{robinson2005topology}
\bibinfo{author}{Robinson, D.} \& \bibinfo{author}{Goforth, D.}
\newblock \emph{\bibinfo{title}{The topology of the 2x2 games: a new periodic table}}, vol.~\bibinfo{volume}{3} (\bibinfo{publisher}{Psychology Press}, \bibinfo{year}{2005}).

\bibitem{jones2008smarter}
\bibinfo{author}{Jones, G.}
\newblock \bibinfo{journal}{\bibinfo{title}{Are smarter groups more cooperative? evidence from prisoner's dilemma experiments, 1959--2003}}.
\newblock {\emph{\JournalTitle{Journal of Economic Behavior \& Organization}}} \textbf{\bibinfo{volume}{68}}, \bibinfo{pages}{489--497} (\bibinfo{year}{2008}).

\bibitem{axelrod1981evolution}
\bibinfo{author}{Axelrod, R.} \& \bibinfo{author}{Hamilton, W.~D.}
\newblock \bibinfo{journal}{\bibinfo{title}{The evolution of cooperation}}.
\newblock {\emph{\JournalTitle{science}}} \textbf{\bibinfo{volume}{211}}, \bibinfo{pages}{1390--1396} (\bibinfo{year}{1981}).

\bibitem{hawkins2016formation}
\bibinfo{author}{Hawkins, R.~X.} \& \bibinfo{author}{Goldstone, R.~L.}
\newblock \bibinfo{journal}{\bibinfo{title}{The formation of social conventions in real-time environments}}.
\newblock {\emph{\JournalTitle{PloS one}}} \textbf{\bibinfo{volume}{11}}, \bibinfo{pages}{e0151670} (\bibinfo{year}{2016}).

\bibitem{young1996economics}
\bibinfo{author}{Young, H.~P.}
\newblock \bibinfo{journal}{\bibinfo{title}{The economics of convention}}.
\newblock {\emph{\JournalTitle{Journal of economic perspectives}}} \textbf{\bibinfo{volume}{10}}, \bibinfo{pages}{105--122} (\bibinfo{year}{1996}).

\bibitem{andalman2004alternation}
\bibinfo{author}{Andalman, A.} \& \bibinfo{author}{Kemp, C.}
\newblock \bibinfo{journal}{\bibinfo{title}{Alternation in the repeated battle of the sexes}}.
\newblock {\emph{\JournalTitle{Cambridge: MIT Press. Andreoni, J., \& Miller, J.(2002). Giving according to GARP: an experimental test of the consistency of preferences for altruism. Econometrica}}} \textbf{\bibinfo{volume}{70}}, \bibinfo{pages}{737753} (\bibinfo{year}{2004}).

\bibitem{lau2008using}
\bibinfo{author}{Lau, S.-H.~P.} \& \bibinfo{author}{Mui, V.-L.}
\newblock \bibinfo{journal}{\bibinfo{title}{Using turn taking to mitigate coordination and conflict problems in the repeated battle of the sexes game}}.
\newblock {\emph{\JournalTitle{Theory and Decision}}} \textbf{\bibinfo{volume}{65}}, \bibinfo{pages}{153--183} (\bibinfo{year}{2008}).

\bibitem{mckelvey2001playing}
\bibinfo{author}{McKelvey, R.~D.} \& \bibinfo{author}{Palfrey, T.~R.}
\newblock \bibinfo{journal}{\bibinfo{title}{Playing in the dark: Information, learning, and coordination in repeated games}}.
\newblock {\emph{\JournalTitle{California Institute of Technology}}}  (\bibinfo{year}{2001}).

\bibitem{arifovic2018learning}
\bibinfo{author}{Arifovic, J.} \& \bibinfo{author}{Ledyard, J.}
\newblock \bibinfo{journal}{\bibinfo{title}{Learning to alternate}}.
\newblock {\emph{\JournalTitle{Experimental Economics}}} \textbf{\bibinfo{volume}{21}}, \bibinfo{pages}{692--721} (\bibinfo{year}{2018}).

\bibitem{swettenham1996s}
\bibinfo{author}{Swettenham, J.}
\newblock \bibinfo{journal}{\bibinfo{title}{What's inside someone's head? conceiving of the mind as a camera helps children with autism acquire an alternative to a theory of mind}}.
\newblock {\emph{\JournalTitle{Cognitive Neuropsychiatry}}} \textbf{\bibinfo{volume}{1}}, \bibinfo{pages}{73--88} (\bibinfo{year}{1996}).

\bibitem{westby2014developmental}
\bibinfo{author}{Westby, C.} \& \bibinfo{author}{Robinson, L.}
\newblock \bibinfo{journal}{\bibinfo{title}{A developmental perspective for promoting theory of mind}}.
\newblock {\emph{\JournalTitle{Topics in language disorders}}} \textbf{\bibinfo{volume}{34}}, \bibinfo{pages}{362--382} (\bibinfo{year}{2014}).

\bibitem{begeer2011theory}
\bibinfo{author}{Begeer, S.} \emph{et~al.}
\newblock \bibinfo{journal}{\bibinfo{title}{Theory of mind training in children with autism: A randomized controlled trial}}.
\newblock {\emph{\JournalTitle{Journal of autism and developmental disorders}}} \textbf{\bibinfo{volume}{41}}, \bibinfo{pages}{997--1006} (\bibinfo{year}{2011}).

\bibitem{moghaddam2023boosting}
\bibinfo{author}{Moghaddam, S.~R.} \& \bibinfo{author}{Honey, C.~J.}
\newblock \bibinfo{journal}{\bibinfo{title}{Boosting theory-of-mind performance in large language models via prompting}}.
\newblock {\emph{\JournalTitle{arXiv preprint arXiv:2304.11490}}}  (\bibinfo{year}{2023}).

\bibitem{ouyang2022training}
\bibinfo{author}{Ouyang, L.} \emph{et~al.}
\newblock \bibinfo{journal}{\bibinfo{title}{Training language models to follow instructions with human feedback}}.
\newblock {\emph{\JournalTitle{Advances in Neural Information Processing Systems}}} \textbf{\bibinfo{volume}{35}}, \bibinfo{pages}{27730--27744} (\bibinfo{year}{2022}).

\bibitem{wolf2023fundamental}
\bibinfo{author}{Wolf, Y.}, \bibinfo{author}{Wies, N.}, \bibinfo{author}{Levine, Y.} \& \bibinfo{author}{Shashua, A.}
\newblock \bibinfo{journal}{\bibinfo{title}{Fundamental limitations of alignment in large language models}}.
\newblock {\emph{\JournalTitle{arXiv preprint arXiv:2304.11082}}}  (\bibinfo{year}{2023}).

\bibitem{ullman2023large}
\bibinfo{author}{Ullman, T.}
\newblock \bibinfo{journal}{\bibinfo{title}{Large language models fail on trivial alterations to theory-of-mind tasks}}.
\newblock {\emph{\JournalTitle{arXiv preprint arXiv:2302.08399}}}  (\bibinfo{year}{2023}).

\bibitem{le2019revisiting}
\bibinfo{author}{Le, M.}, \bibinfo{author}{Boureau, Y.-L.} \& \bibinfo{author}{Nickel, M.}
\newblock \bibinfo{title}{Revisiting the evaluation of theory of mind through question answering}.
\newblock In \emph{\bibinfo{booktitle}{Proceedings of the 2019 Conference on Empirical Methods in Natural Language Processing and the 9th International Joint Conference on Natural Language Processing (EMNLP-IJCNLP)}}, \bibinfo{pages}{5872--5877} (\bibinfo{year}{2019}).

\bibitem{kosinski2023theory}
\bibinfo{author}{Kosinski, M.}
\newblock \bibinfo{journal}{\bibinfo{title}{Theory of mind may have spontaneously emerged in large language models}}.
\newblock {\emph{\JournalTitle{arXiv preprint arXiv:2302.02083}}} \textbf{\bibinfo{volume}{4}}, \bibinfo{pages}{169} (\bibinfo{year}{2023}).

\bibitem{horton2023large}
\bibinfo{author}{Horton, J.~J.}
\newblock \bibinfo{title}{Large language models as simulated economic agents: What can we learn from homo silicus?}
\newblock \bibinfo{type}{Tech. Rep.}, \bibinfo{institution}{National Bureau of Economic Research} (\bibinfo{year}{2023}).

\bibitem{aher2022using}
\bibinfo{author}{Aher, G.~V.}, \bibinfo{author}{Arriaga, R.~I.} \& \bibinfo{author}{Kalai, A.~T.}
\newblock \bibinfo{title}{Using large language models to simulate multiple humans and replicate human subject studies}.
\newblock In \emph{\bibinfo{booktitle}{International Conference on Machine Learning}}, \bibinfo{pages}{337--371} (\bibinfo{organization}{PMLR}, \bibinfo{year}{2023}).

\bibitem{dalbo2011evolution}
\bibinfo{author}{Dal B\'o, P.} \& \bibinfo{author}{Fr\'echette, G.~R.}
\newblock \bibinfo{journal}{\bibinfo{title}{The evolution of cooperation in infinitely repeated games: Experimental evidence}}.
\newblock {\emph{\JournalTitle{Am. Econ. Rev.}}} \textbf{\bibinfo{volume}{101}}, \bibinfo{pages}{411--429} (\bibinfo{year}{2011}).

\bibitem{nowak2005evolution}
\bibinfo{author}{Nowak, M.~A.} \& \bibinfo{author}{Sigmund, K.}
\newblock \bibinfo{journal}{\bibinfo{title}{Evolution of indirect reciprocity}}.
\newblock {\emph{\JournalTitle{Nature}}} \textbf{\bibinfo{volume}{437}}, \bibinfo{pages}{1291--1298} (\bibinfo{year}{2005}).

\bibitem{radford2019language}
\bibinfo{author}{Radford, A.} \emph{et al.}
\newblock \bibinfo{journal}{\bibinfo{title}{Language models are unsupervised multitask learners}}.
\newblock {\emph{\JournalTitle{OpenAI blog}}} \textbf{\bibinfo{volume}{1}}, \bibinfo{pages}{9} (\bibinfo{year}{2019}).

\bibitem{nowak2006five}
\bibinfo{author}{Nowak, M.~A.}
\newblock \bibinfo{journal}{\bibinfo{title}{Five rules for the evolution of cooperation}}.
\newblock {\emph{\JournalTitle{Science}}} \textbf{\bibinfo{volume}{314}}, \bibinfo{pages}{1560--1563} (\bibinfo{year}{2006}).

\bibitem{wei2022chain}
\bibinfo{author}{Wei, J.} \emph{et~al.}
\newblock \bibinfo{journal}{\bibinfo{title}{Chain-of-thought prompting elicits reasoning in large language models}}.
\newblock {\emph{\JournalTitle{Advances in neural information processing systems}}} \textbf{\bibinfo{volume}{35}}, \bibinfo{pages}{24824--24837} (\bibinfo{year}{2022}).

\bibitem{engle2004evolution}
\bibinfo{author}{Engle-Warnick, J.} \& \bibinfo{author}{Slonim, R.~L.}
\newblock \bibinfo{journal}{\bibinfo{title}{The evolution of strategies in a repeated trust game}}.
\newblock {\emph{\JournalTitle{Journal of Economic Behavior \& Organization}}} \textbf{\bibinfo{volume}{55}}, \bibinfo{pages}{553--573} (\bibinfo{year}{2004}).

\bibitem{rankin2007tragedy}
\bibinfo{author}{Rankin, D.~J.}, \bibinfo{author}{Bargum, K.} \& \bibinfo{author}{Kokko, H.}
\newblock \bibinfo{journal}{\bibinfo{title}{The tragedy of the commons in evolutionary biology}}.
\newblock {\emph{\JournalTitle{Trends in ecology \& evolution}}} \textbf{\bibinfo{volume}{22}}, \bibinfo{pages}{643--651} (\bibinfo{year}{2007}).

\bibitem{rahwan2022machine}
\bibinfo{author}{Rahwan, I.} \emph{et~al.}
\newblock \bibinfo{journal}{\bibinfo{title}{Machine behaviour}}.
\newblock {\emph{\JournalTitle{Machine Learning and the City: Applications in Architecture and Urban Design}}} \bibinfo{pages}{143--166} (\bibinfo{year}{2022}).

\bibitem{schulz2020computational}
\bibinfo{author}{Schulz, E.} \& \bibinfo{author}{Dayan, P.}
\newblock \bibinfo{journal}{\bibinfo{title}{Computational psychiatry for computers}}.
\newblock {\emph{\JournalTitle{Iscience}}} \textbf{\bibinfo{volume}{23}}, \bibinfo{pages}{101772} (\bibinfo{year}{2020}).

\bibitem{binz2023using}
\bibinfo{author}{Binz, M.} \& \bibinfo{author}{Schulz, E.}
\newblock \bibinfo{journal}{\bibinfo{title}{Using cognitive psychology to understand gpt-3}}.
\newblock {\emph{\JournalTitle{Proceedings of the National Academy of Sciences}}} \textbf{\bibinfo{volume}{120}}, \bibinfo{pages}{e2218523120} (\bibinfo{year}{2023}).

\bibitem{coda2023inducing}
\bibinfo{author}{Coda-Forno, J.} \emph{et~al.}
\newblock \bibinfo{journal}{\bibinfo{title}{Inducing anxiety in large language models increases exploration and bias}}.
\newblock {\emph{\JournalTitle{arXiv preprint arXiv:2304.11111}}}  (\bibinfo{year}{2023}).

\bibitem{kojima2022large}
\bibinfo{author}{Kojima, T.}, \bibinfo{author}{Gu, S.~S.}, \bibinfo{author}{Reid, M.}, \bibinfo{author}{Matsuo, Y.} \& \bibinfo{author}{Iwasawa, Y.}
\newblock \bibinfo{journal}{\bibinfo{title}{Large language models are zero-shot reasoners}}.
\newblock {\emph{\JournalTitle{Advances in neural information processing systems}}} \textbf{\bibinfo{volume}{35}}, \bibinfo{pages}{22199--22213} (\bibinfo{year}{2022}).

\bibitem{rabinowitz2018machine}
\bibinfo{author}{Rabinowitz, N.} \emph{et~al.}
\newblock \bibinfo{title}{Machine theory of mind}.
\newblock In \emph{\bibinfo{booktitle}{International conference on machine learning}}, \bibinfo{pages}{4218--4227} (\bibinfo{organization}{PMLR}, \bibinfo{year}{2018}).

\bibitem{cuzzolin2020knowing}
\bibinfo{author}{Cuzzolin, F.}, \bibinfo{author}{Morelli, A.}, \bibinfo{author}{Cirstea, B.} \& \bibinfo{author}{Sahakian, B.~J.}
\newblock \bibinfo{journal}{\bibinfo{title}{Knowing me, knowing you: theory of mind in ai}}.
\newblock {\emph{\JournalTitle{Psychological medicine}}} \textbf{\bibinfo{volume}{50}}, \bibinfo{pages}{1057--1061} (\bibinfo{year}{2020}).

\bibitem{alon2022dis}
\bibinfo{author}{Alon, N.}, \bibinfo{author}{Schulz, L.}, \bibinfo{author}{Dayan, P.} \& \bibinfo{author}{Rosenschein, J.}
\newblock \bibinfo{title}{A (dis-) information theory of revealed and unrevealed preferences}.
\newblock In \emph{\bibinfo{booktitle}{NeurIPS 2022 Workshop on Information-Theoretic Principles in Cognitive Systems}} (\bibinfo{year}{2022}).

\bibitem{frith2005theory}
\bibinfo{author}{Frith, C.} \& \bibinfo{author}{Frith, U.}
\newblock \bibinfo{journal}{\bibinfo{title}{Theory of mind}}.
\newblock {\emph{\JournalTitle{Current biology}}} \textbf{\bibinfo{volume}{15}}, \bibinfo{pages}{R644--R645} (\bibinfo{year}{2005}).

\bibitem{velez2021learning}
\bibinfo{author}{V{\'e}lez, N.} \& \bibinfo{author}{Gweon, H.}
\newblock \bibinfo{journal}{\bibinfo{title}{Learning from other minds: An optimistic critique of reinforcement learning models of social learning}}.
\newblock {\emph{\JournalTitle{Current opinion in behavioral sciences}}} \textbf{\bibinfo{volume}{38}}, \bibinfo{pages}{110--115} (\bibinfo{year}{2021}).

\bibitem{lissek2008cooperation}
\bibinfo{author}{Lissek, S.} \emph{et~al.}
\newblock \bibinfo{journal}{\bibinfo{title}{Cooperation and deception recruit different subsets of the theory-of-mind network}}.
\newblock {\emph{\JournalTitle{PloS one}}} \textbf{\bibinfo{volume}{3}}, \bibinfo{pages}{e2023} (\bibinfo{year}{2008}).

\bibitem{hula2015monte}
\bibinfo{author}{Hula, A.}, \bibinfo{author}{Montague, P.~R.} \& \bibinfo{author}{Dayan, P.}
\newblock \bibinfo{journal}{\bibinfo{title}{Monte carlo planning method estimates planning horizons during interactive social exchange}}.
\newblock {\emph{\JournalTitle{PLoS computational biology}}} \textbf{\bibinfo{volume}{11}}, \bibinfo{pages}{e1004254} (\bibinfo{year}{2015}).

\bibitem{ho2022planning}
\bibinfo{author}{Ho, M.~K.}, \bibinfo{author}{Saxe, R.} \& \bibinfo{author}{Cushman, F.}
\newblock \bibinfo{journal}{\bibinfo{title}{Planning with theory of mind}}.
\newblock {\emph{\JournalTitle{Trends in Cognitive Sciences}}}  (\bibinfo{year}{2022}).

\bibitem{han2021or}
\bibinfo{author}{Han, T.~A.}, \bibinfo{author}{Perret, C.} \& \bibinfo{author}{Powers, S.~T.}
\newblock \bibinfo{journal}{\bibinfo{title}{When to (or not to) trust intelligent machines: Insights from an evolutionary game theory analysis of trust in repeated games}}.
\newblock {\emph{\JournalTitle{Cognitive Systems Research}}} \textbf{\bibinfo{volume}{68}}, \bibinfo{pages}{111--124} (\bibinfo{year}{2021}).

\bibitem{chan2023towards}
\bibinfo{author}{Chan, A.}, \bibinfo{author}{Rich{\'e}, M.} \& \bibinfo{author}{Clifton, J.}
\newblock \bibinfo{journal}{\bibinfo{title}{Towards the scalable evaluation of cooperativeness in language models}}.
\newblock {\emph{\JournalTitle{arXiv preprint arXiv:2303.13360}}}  (\bibinfo{year}{2023}).

\bibitem{dasgupta2022language}
\bibinfo{author}{Dasgupta, I.} \emph{et~al.}
\newblock \bibinfo{journal}{\bibinfo{title}{Language models show human-like content effects on reasoning}}.
\newblock {\emph{\JournalTitle{arXiv preprint arXiv:2207.07051}}}  (\bibinfo{year}{2022}).

\bibitem{crandall2011learning}
\bibinfo{author}{Crandall, J.~W.} \& \bibinfo{author}{Goodrich, M.~A.}
\newblock \bibinfo{journal}{\bibinfo{title}{Learning to compete, coordinate, and cooperate in repeated games using reinforcement learning}}.
\newblock {\emph{\JournalTitle{Machine Learning}}} \textbf{\bibinfo{volume}{82}}, \bibinfo{pages}{281--314} (\bibinfo{year}{2011}).

\bibitem{goodfellow2020generative}
\bibinfo{author}{Goodfellow, I.} \emph{et~al.}
\newblock \bibinfo{journal}{\bibinfo{title}{Generative adversarial networks}}.
\newblock {\emph{\JournalTitle{Communications of the ACM}}} \textbf{\bibinfo{volume}{63}}, \bibinfo{pages}{139--144} (\bibinfo{year}{2020}).

\bibitem{santos2024prosocial}
\bibinfo{author}{Santos, F.~P.}
\newblock \bibinfo{journal}{\bibinfo{title}{Prosocial dynamics in multiagent systems}}.
\newblock {\emph{\JournalTitle{AI Magazine}}} \textbf{\bibinfo{volume}{45}}, \bibinfo{pages}{131--138} (\bibinfo{year}{2024}).

\bibitem{guo2023facilitating}
\bibinfo{author}{Guo, H.} \emph{et~al.}
\newblock \bibinfo{journal}{\bibinfo{title}{Facilitating cooperation in human-agent hybrid populations through autonomous agents}}.
\newblock {\emph{\JournalTitle{Iscience}}} \textbf{\bibinfo{volume}{26}} (\bibinfo{year}{2023}).

\bibitem{powers2023stuff}
\bibinfo{author}{Powers, S.~T.} \emph{et~al.}
\newblock \bibinfo{journal}{\bibinfo{title}{The stuff we swim in: regulation alone will not lead to justifiable trust in ai}}.
\newblock {\emph{\JournalTitle{IEEE Technology and Society Magazine}}} \textbf{\bibinfo{volume}{42}}, \bibinfo{pages}{95--106} (\bibinfo{year}{2023}).

\bibitem{palan2018prolific}
\bibinfo{author}{Palan, S.} \& \bibinfo{author}{Schitter, C.}
\newblock \bibinfo{journal}{\bibinfo{title}{Prolific. ac—a subject pool for online experiments}}.
\newblock {\emph{\JournalTitle{J. behavioral experimental finance}}} \textbf{\bibinfo{volume}{17}}, \bibinfo{pages}{22--27} (\bibinfo{year}{2018}).

\bibitem{normann2012impact}
\bibinfo{author}{Normann, H.-T.} \& \bibinfo{author}{Wallace, B.}
\newblock \bibinfo{journal}{\bibinfo{title}{The impact of the termination rule on cooperation in a prisoner’s dilemma experiment}}.
\newblock {\emph{\JournalTitle{Int. J. Game Theory}}} \textbf{\bibinfo{volume}{41}}, \bibinfo{pages}{707--718} (\bibinfo{year}{2012}).

\bibitem{charness2002understanding}
\bibinfo{author}{Charness, G.} \& \bibinfo{author}{Rabin, M.}
\newblock \bibinfo{journal}{\bibinfo{title}{Understanding social preferences with simple tests}}.
\newblock {\emph{\JournalTitle{The quarterly journal economics}}} \textbf{\bibinfo{volume}{117}}, \bibinfo{pages}{817--869} (\bibinfo{year}{2002}).

\bibitem{wong2005dynamic}
\bibinfo{author}{Wong, R.~Y.-m.} \& \bibinfo{author}{Hong, Y.-y.}
\newblock \bibinfo{journal}{\bibinfo{title}{Dynamic influences of culture on cooperation in the prisoner’s dilemma}}.
\newblock {\emph{\JournalTitle{Psychol. science}}} \textbf{\bibinfo{volume}{16}}, \bibinfo{pages}{429--434} (\bibinfo{year}{2005}).

\bibitem{liu2023pre}
\bibinfo{author}{Liu, P.} \emph{et~al.}
\newblock \bibinfo{journal}{\bibinfo{title}{Pre-train, prompt, and predict: A systematic survey of prompting methods in natural language processing}}.
\newblock {\emph{\JournalTitle{ACM Computing Surveys}}} \textbf{\bibinfo{volume}{55}}, \bibinfo{pages}{1--35} (\bibinfo{year}{2023}).

\bibitem{rouder2009bayesian}
\bibinfo{author}{Rouder, J.~N.}, \bibinfo{author}{Speckman, P.~L.}, \bibinfo{author}{Sun, D.}, \bibinfo{author}{Morey, R.~D.} \& \bibinfo{author}{Iverson, G.}
\newblock \bibinfo{journal}{\bibinfo{title}{Bayesian t tests for accepting and rejecting the null hypothesis}}.
\newblock {\emph{\JournalTitle{Psychonomic bulletin \& review}}} \textbf{\bibinfo{volume}{16}}, \bibinfo{pages}{225--237} (\bibinfo{year}{2009}).

\end{thebibliography}
\end{document}